\documentclass[journal]{IEEEtran}
\usepackage{graphicx}
\usepackage{subfig}
\usepackage{amsmath, amsfonts}
\usepackage{amssymb}
\usepackage{url}
\usepackage{hyperref}
\usepackage{cite}
\usepackage{multirow}
\usepackage{booktabs}
\usepackage{paralist}
\usepackage{booktabs}
\usepackage{xcolor}
\usepackage{soul}
\usepackage{gensymb}
\usepackage{textcomp}

\usepackage{pgfplots}
\pgfplotsset{width=8.5cm, compat=1.9}

\graphicspath{{./figs/}}

\begin{document}

\title{An Interpretable Multiple-Instance Approach for the Detection of 
  referable Diabetic Retinopathy from Fundus Images}

\author{
  Alexandros Papadopoulos$^1$,~\IEEEmembership{Student~Member,~IEEE,}
  Fotis Topouzis$^2$,
  Anastasios Delopoulos$^1$,~\IEEEmembership{Member,~IEEE,}
  \thanks{$^1$Multimedia Understanding Group, Information Processing Laboratory, Dept. of Electrical and Computer Engineering, Aristotle University of Thessaloniki, Greece
  $^2$Department of Ophthalmology, Aristotle University Medical School, Thessaloniki, Greece}
}

\maketitle

\begin{abstract}
 Diabetic Retinopathy (DR) is a leading cause of vision loss globally. 
 Yet despite its prevalence, the majority of affected people lack access to the specialized
 ophthalmologists and equipment required for assessing their condition. 
 This can lead to delays in the start of treatment, thereby lowering their chances
 for a successful outcome.
 Machine learning systems that automatically detect the disease in eye fundus images 
 have been proposed as a means of facilitating access to DR
 severity estimates for patients in remote regions or even for complementing the human
 expert's diagnosis. In this paper, we propose a machine learning system for the
 detection of referable DR in fundus images that is based on the paradigm of multiple-instance
 learning. By extracting local information from image patches and 
 combining it efficiently through an attention mechanism,
 our system is able to achieve high classification accuracy. 
 Moreover, it can
 highlight potential image regions where DR manifests through its characteristic lesions.
 We evaluate our approach on publicly available retinal image datasets, in which it
 exhibits near state-of-the-art performance, while also
 producing interpretable visualizations of its predictions.

\end{abstract}

\IEEEpeerreviewmaketitle

\section{Introduction}
\label{sec:introduction}
\IEEEPARstart{D}{iabetic} Retinopathy (DR) is a complication of 
diabetes mellitus that can lead to blindness
if left untreated. More than 1 out of 3 diabetic patients are expected to develop DR 
during their lifetime \cite{congdon2003important}, with their chances
increasing over time \cite{Yau556}. Yet, despite its prevalence among
diabetic populations, the risk of blindness can be 
significantly reduced via timely treatment, i.e.
before the retina is severely damaged \cite{Ciulla2653}. 

Diabetic Retinopathy is characterized as either non-proliferative (NPDR), meaning it 
manifests mainly through retinal lesions,
or as proliferative (PDR), meaning neovascularization of 
weak blood vessels also occurs. 
The International Clinical
Diabetic Retinopathy Disease (ICDR) severity scale \cite{wu2013classification}, 
suggests a finer classification of the disease into the following 5 
stages, based on the observable
findings during eye examination:
\begin{inparaenum}
	\item No apparent retinopathy
	\item Mild NPDR
	\item Moderate NPDR
	\item Severe NPDR
	\item PDR.
\end{inparaenum}
Common guidelines recommend annual screenings for diabetic patients without
or with mild DR, 6 month follow up examination for
moderate DR, and referral to an ophthalmologist for treatment evaluation 
for severe cases \cite{chakrabarti2012diabetic}. 
A DR of moderate or worse stage is further characterized as
\emph{referable Diabetic Retinopathy (rDR)}.
Subjects who were diagnosed with rDR by a trained DR grader
(not necessarily an ophthalmologist)
must also be referred to an ophthalmologist 
for further evaluation of their condition.

Screening for DR is usually carried out by either an in-person eye examination or by means of 
retinal photography. In either case, an eye care professional examines the
retina (either directly through a slit-lamp or 
indirectly through a high resolution retinal photograph
captured with a specialized camera) for signs of 
the disease, such as microaneurysms, haemorrhages
and hard or soft exudates. Other factors such as macular edema,
narrowing of the blood vessels or damage in the nerve 
tissue are also considered \cite{Ciulla2653}. In general, 
accurate DR grading is a daunting task even for experienced graders, and,
as a result, inter-grader variability is quite common \cite{krause2018grader}.

In recent years, retinal photography has been widely
accepted as an adequate screening method that can even lead to improved
diagnostic performance compared to the standard
slit-lamp examination \cite{olson2003comparative}. 
This acceptance has naturally led to the conduction of much research
on the development of automated methods for grading retinal images,
as such techniques can provide
substantial benefits to the standard DR
screening procedure. For example, they can support retinal 
specialists by lightening their workload or by identifying cases
they might have missed. Automated grading can also be used to increase
the coverage of nation-wide screening programs by facilitating
access for people in remote or rural regions via the use of teleophthalmology. 

Image classification is a classic use case for machine learning (ML)
algorithms. especially those based on the paradigm of deep learning.
In recent years, deep learning has achieved impressive results
in a series of tasks, including image classification. 
In this work, we propose a method based on deep learning
to detect referable Diabetic Retinopathy from retinal images and 
simultaneously produce heatmaps of the most dominant DR lesions to 
aid the model's interpretability. 
Our method operates by independently extracting information from
multiple small image patches and combining them based on their content.
We focus on the detection of rDR,
as this simplified binary classification task is a common target 
for automated DR grading methods that
provides ample clinical utility \cite{abramoff2013automated}.

Our paper is organized in the following way: in 
Section \ref{sec:soa} we review
the most prominent works in the related literature. 
Then, in Section \ref{sec:method} we introduce
the proposed methodology for performing rDR detection. In Section
\ref{sec:datasets} we provide details about the datasets used in
our experiments and in Section \ref{sec:experiments} we present
the experimental results for our method, as well as comparisons with 
other works in the literature. Finally, we discuss our method and results
in Section \ref{sec:discussion} and conclude the paper
with Section \ref{sec:conclusion}.

\section{Related work} \label{sec:soa}
The automated analysis of retinal images for diabetic retinopathy detection
has received impressive research attention over the last decade, particularly
following the rise of deep learning that eliminated the need of manually
constructing problem-specific features. Many different learning tasks
have been pursued in the related literature, such as 
optic disk and blood vessel segmentation, lesion detection and 
DR grading.
As the relevant literature is abundant, in this section we will 
focus on the tasks of DR grading and lesion detection that are mostly 
related to our work. For a thorough review of the field, we refer
the reader to the excellent survey of \cite{stolte2020survey}.

In grading approaches, the task is usually cast as a binary 
classification problem of no rDR vs 
rDR, but it can be also viewed as a multi-class problem, using the 
stages of the ICDR severity scale as 5 distinct classes,
Early methods relied on 
traditional computer vision methods, mainly using feature extraction
techniques to identify specific properties of interest in the fundus image,
coupled with shallow machine learning models for classification.
For instance, \cite{nayak2008automated} used morphological and 
texture analysis to extract blood vessel and hard exudate features
which were then used
as input features in a neural network, 
\cite{acharya2008application} made use of higher-order spectra coupled with an
SVM classifier and, more recently, \cite{seoud2015red} proposed a set of high-performing,
handcrafted shape features that were fed to a Random Forest classifier.

In recent years, the rise of \emph{Deep Convolutional Neural Networks (DCNN)}
and their dominance 
in computer vision tasks has transformed the research landscape,
so that most works are nowadays based on deep neural networks.
The trend reached its peak with the seminal work \cite{gulshan2016development} 
that reported excellent rDR classification performance (ROC AUC of 0.99), 
by using an Inception v3 model pre-trained on ImageNet and finetuned on
a private dataset of 120k fundus images. 
A replication study using the publicly-available
Kaggle-EyePACS dataset \cite{graham2015kaggle}, was conducted 
in \cite{voets2018replication}, but was unable to reproduce
the results of the original study, suggesting that the ground-truth
quality is a decisive factor for achieving high performance, as
the dataset used in \cite{gulshan2016development} 
was annotated by a panel of 7 retina specialists,
while Kaggle-EyePACS was annotated by a single specialist.
This is further corroborated by the findings of \cite{krause2018grader},
who report performance improvements when using even a small
amount of adjudicated DR ground truths.

The work of \cite{pires2019data} investigates the use of a multi-resolution
training scheme, using different networks with shared weights for each
different resolution of the input image. They also extract and combine
features simultaneously from multiple label-preserving 
random perturbations of the same input image, 
as suggested by the team that placed 
second\footnote{https://www.kaggle.com/c/diabetic-retinopathy-detection/forums/t/
	15617/team-o-o-solution-summary} in the Kaggle DR competition, along with other
tricks, such as aggressive data augmentation, to improve performance.

An interesting addition to the standard DR prediction procedure
is suggested by the authors of \cite{leibig2017leveraging}. In particular, they
incorporate a bayesian estimation of the model's
uncertainty during test time, implemented via a dropout 
approximation \cite{gal2016dropout}. They go on to show,
that by refraining from classifying images for which the model is uncertain,
one can increase classification performance and obtain more reliable predictions.
This modification is applicable to any of the usual deep CNN models 
that have been suggested in the context of DR classification, such as 
ResNet \cite{gargeya2017automated}, AlexNet \cite{abramoff2013automated} 
and VGG \cite{pires2019data}.

A different strategy 
that is more on par with the screening procedure of eye specialists is 
to first detect specific DR lesions and then use these detections
to infer DR. Following this direction, \cite{abramoff2013automated}
introduced a series of lesion detector models, based on
AlexNet \cite{krizhevsky2012imagenet} and  VGG \cite{simonyan2014very},
where each detector is applied to the input image to detect a particular type
of DR lesion (haemorrhages, exudates, etc). The detector outputs 
are then fused together to form a feature vector, that is used to
first assess the image quality, and then, if the quality
is found acceptable, to predict the level of DR. The resulting hybrid system
was shown to achieve very high performance and has been deployed in
real world conditions. 

Such methods, however, depend on the manual annotation
of DR lesions, a procedure that significantly increases the clinician workload 
and, as such, can be typically be carried out for few images. In addition, 
disease-relevant information tend to be located in just a few regions in the input image
\cite{kandemir2015computer}, so that using a single feature vector
to represent the entire image can be problematic. 
To aid in these problems, \emph{Multiple-Instance Learning} methods \cite{maron1998framework} that
treat the input image as bag of patches accompanied by a single DR label,
have been suggested. An early MIL method introduced in \cite{quellec2012multiple}, 
used rectangular patches originating from DR-positive images to train a model
using a patch relevance criterion. The resulting model was then shown to produce 
high relevance scores for patches that contained DR lesions.
The subsequent work of \cite{kandemir2015computer} employed the use of generic 
MIL algorithms like the mi-SVM method \cite{andrews2003support}, to produce both local (per-patch)
and global (per-image) decisions, by processing each patch independently and
combining the per-patch results according to some aggregation criterion. A similar
method that made use of deep learning was later proposed in \cite{zhou2017deep}.



\begin{figure*}[!ht]
	\centering
	\includegraphics[width=1\linewidth]{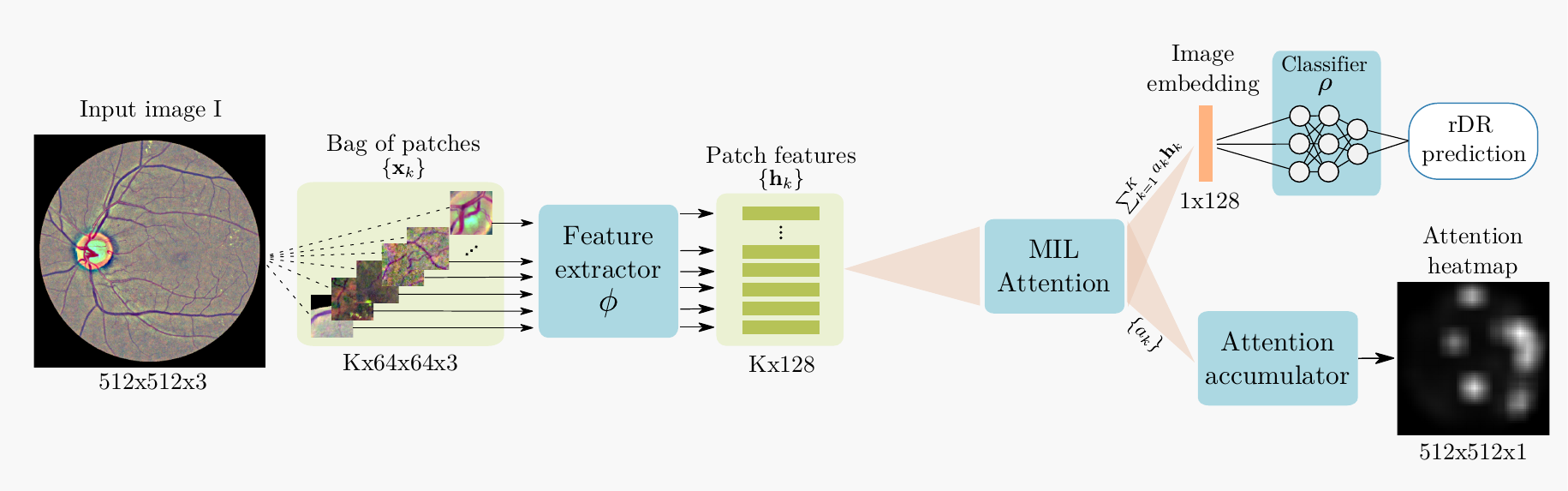}
	\caption{High-level overview of the proposed rDR prediction pipeline. 
		The preprocessed image $I$ is represented by K rectangular patches 
		with fixed overlap that cover most of the
		retinal disk (patches that have less than 50\% retinal content, such as
		the patches extracted near the boundary of the retinal disk, are discarded). 
                Features are then extracted from each $\mathbf{x}_k$ patch via a Resnet-18 (function $\phi$) model 
                Each resulting feature vector $\mathbf{h}_k$ is assigned a weight $\alpha_k$
                by an attention mechanism and the weighted average of all the vectors $\{\mathbf{h}_k\}$
                is used to acquire a single embedding that describes the entire image.
                Finally, the embedding is fed to a fully connected 
                layer (function $\rho$) that outputs the probability of rDR for the given image. At the same time,
                the attention weights of different patches are mapped to their corresponding
                image regions and an attention heatmap that highlights the most
                salient image regions according to the model is constructed.
	}
	\label{pipeline}
\end{figure*}

Here we introduce a method to perform rDR classification
by analyzing the fundus image on the patch level. Our approach
follows the MIL paradigm that treats each input image as a bag of 
image patches. At first, each patch is independently encoded via a deep CNN model
into a feature vector. Contrary to other MIL approaches in which the individual
patch vectors are pooled via pre-defined operators, here
we examine the use of the attention mechanism \cite{pmlr-v80-ilse18a}. 
In the context of MIL, attention serves as a trainable pooling operator
that learns to put emphasis on the most informative 
instances (i.e. patches that contain DR lesions) and ignore the
uninformative ones, thus leading
to an image-level representation that retains the most relevant
information and can thus be used for efficient rDR detection. 

In medical applications, model interpretability is 
key for the successful adoption of a proposed ML solution, as
human experts are more likely to trust a decision if they
understand the motives behind it. In the case of DR in particular, a model
is usually considered black-box during training and endowed with intrepretability
properties using some post-training optimization procedure to
output a heatmap of the fundus regions 
\cite{gargeya2017automated, quellec2017deep, wang2017zoom} 
that the model is most sensitive to.
In this work, we follow a different approach: we leverage the 
attention mechanism to construct
heatmaps that highlight the image regions that directly 
contributed to a given prediction. In doing so, heatmap generation
arises quite naturally since attention is a basic building block of our model.


We evaluate the classification performance of our model on the publicly available
Kaggle-EyePACS \cite{dataset} and Messidor-2 \cite{abramoff2013automated} datasets.
In the rDR detection task, an ensemble of 5 models 
trained on the standard Kaggle-EyePACS training set,
achieves a ROC AUC of $0.957$ on the Kaggle-EyePACS test set and $0.976$ on Messidor-2,
outperforming other patch-based methods and performing on-par with the state-of-the-art.
We also perform experiments to evaluate the 
quality of the produced attention 
heatmaps using the images from the IDRiD 
\cite{h25w98-18} dataset that contain detailed pixel-level
lesion annotations. Our results in these experiments show 
that the attention mechanism
can efficiently recognize the DR-induced lesions in the
fundus of the eye.

\section{Proposed methodology}\label{sec:method}

\subsection{Image pre-processing}\label{sec:preproc}
A practical DR grading system must be able to handle fundus images 
of different resolutions, lighting 
conditions and view points of the retina disk, as these
parameters may vary between different retinal cameras and
capturing conditions.
However, typical image classification models
operate on images of pre-defined dimensions. In addition, 
artificial artifacts in the image, such as blobs caused by dust
in the lens or by improper lightning, can seriously hinder model performance.
Therefore, to mitigate these issues we apply a common pre-processing step 
to all images prior to any machine learning analysis. 
The pre-processing pipeline consists of the following steps:
\begin{enumerate}
	\item Estimate the center and radius 
	of the circular eye disk by means of the Hough transform \footnote{Can be skipped if ROIs are provided by the retinal camera}. 
	\item Crop a rectangular region with edge equal 
	to the estimated eye disk radius and centered at the estimated disk center. 
	\item Resize the rectangular crop to a common resolution of $512\times512$.
        \item Subtract the local color average as suggested in \cite{graham2015kaggle}, to
          account for the variability in lightning conditions across images.
        \item Zero-out the outer 5\% of the retina disk to
          remove artificial boundary effects introduced by the filtering procedure.
\end{enumerate}

\begin{figure}[!h]
	\captionsetup[subfloat]{labelformat=empty}
	\centering
	\subfloat[]{
		\includegraphics[scale=0.5,width=0.4\linewidth]{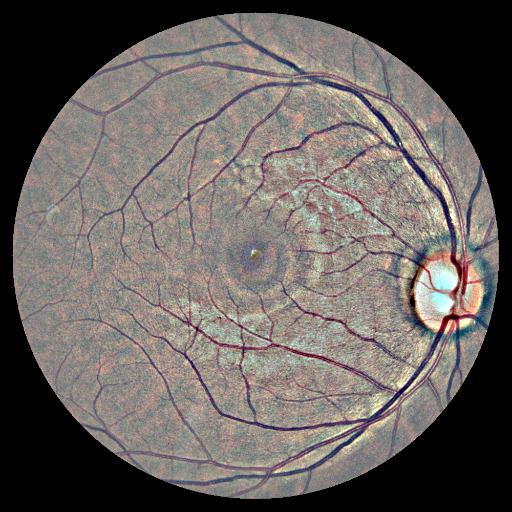}
	}
	\subfloat[]{
		\includegraphics[scale=0.5,width=0.4\linewidth]{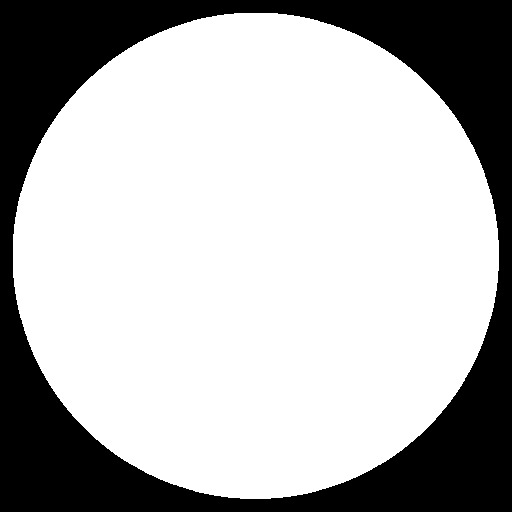}
	}
	\hspace{0mm}
	\subfloat[]{
		\includegraphics[scale=0.5,width=0.4\linewidth]{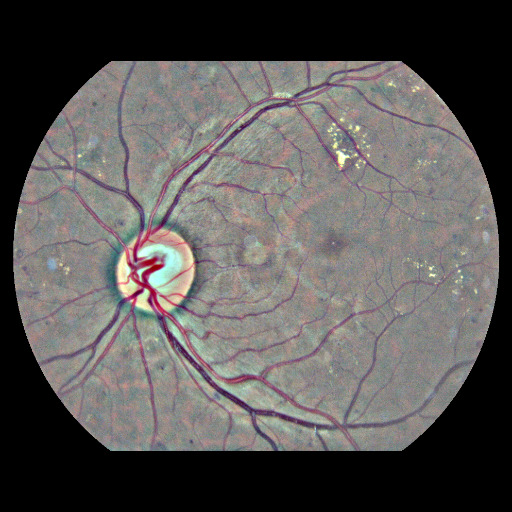}
	}
	\subfloat[]{
		\includegraphics[scale=0.5,width=0.4\linewidth]{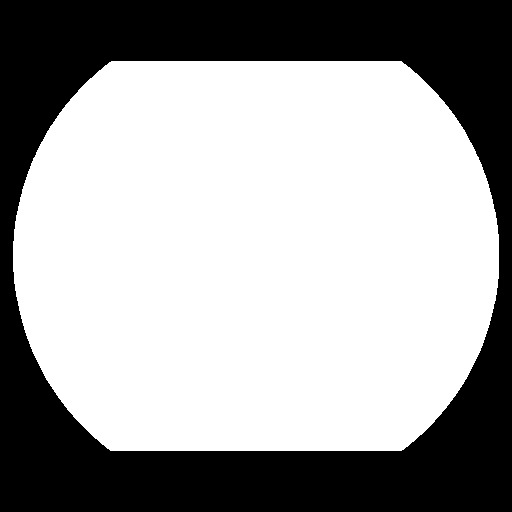}
	}
	
	\caption{Samples of preprocessed retinal images (left column) along with their 
		corresponding binary retina masks (right column).}\label{preprocsamples}
\end{figure}

Images for which the disk center cannot be found via the Hough transform are
discarded altogether.
Additional care is taken for images that
do not contain the whole circular disk (for example \ref{cut1}, \ref{cut2}). 
In such cases, an additional zero
padding step of appropriate size is applied along the height dimension,
before cropping the rectangular retina region.
The overall process results into two components that 
will used in the subsequent learning pipeline: the preprocessed image,
which is ready to be fed to a machine learning model, and the
binary retina mask (obtained during the Hough transform step) 
which can be used to separate the retina from the image background. 
Samples of preprocessed images and their corresponding retina masks are 
given in Fig. \ref{preprocsamples}.

\subsection{Bag of patches encoding}\label{sec:bop}
Having established a common frame of reference for all images, we 
now compute the bag of patches representation for an image. To this end,
rectangular patches of size $d\times d$ pixels are extracted from 
each image. We use different policies for patch extraction during
training and testing. During training
we randomly select $K_{o}$ patches from the pool of possible image patches.
This significantly speeds up training as we can represent an image
using only a small subset of all the possible patches and also adds a regularization
effect, as the representation of a specific image will be different each time
it is fed to the network (since it will be represented by 
a set of different random $K_{o}$ patches).
At test time, we perform exhaustive patch extraction over a regular grid with a 
patch overlap ratio of $t\in[0,1)$.
In both cases, some patches will contain very small part of 
the retina or originate completely 
from the image background. To avoid such cases, we use the previously
computed retina mask and discard patches whose eye disk content is less than $50\%$. 
The surviving $K$ patches form a set $X=\{\mathbf{x}_1,
\mathbf{x}_2, \dots, \mathbf{x}_K\}$ with $\mathbf{x}_k \in \mathbb{R}^{d \times d \times 3}$ 
that constitutes the bag of patches for
the given image. In the following, we will discuss
how we can efficiently utilize the bag of patches
representation to perform rDR prediction. 

\subsection{Multiple-Instance modelling} \label{sec:mil}

In the MIL view of the rDR detection problem, we are 
essentially interested in learning a
function $f$ that will map any set $X=\{\mathbf{x}_1,
\mathbf{x}_2, \dots, \mathbf{x}_K\}$ with $\mathbf{x}_k \in \mathbb{R}^N$ 
(corresponding to the bag of patches of the input image) 
to a real number
(corresponding to the probability of rDR for that image). 
Being a set function, $f$ must be invariant to the different permutations of $X$.
Previous theoretical work \cite{NIPS2017_6931} suggested that any permutation-invariant
function over a set $X$ can be modelled as a sum decomposition 
of the form
$\rho \left( \mathop{\sum}_{\mathbf{x} \in X}\phi\left(\mathbf{x}
\right) \right)$, where the transformation $\phi:\mathbb{R}^{N} \mapsto \mathbb{R}^M$
is applied elementwise
to each set instance, thus leading to a transformed set 
$H= \{\mathbf{h}_1, \mathbf{h}_2, \dots, \mathbf{h}_K\} = \{\phi(\mathbf{x}_1), \phi(\mathbf{x}_2), \dots, \phi(\mathbf{x}_K),\}$,
while the transformation $\rho:\mathbb{R}^M \mapsto \mathbb{R}$
is applied to the result of summing the elements of $H$ to obtain the desired output.
Stepping on this result, \cite{pmlr-v80-ilse18a} proposed
to incorporate the attention-mechanism \cite{NIPS2017_7181} in
the sum-decomposition, as an elegant way of tackling multiple-instance classification 
problems.
More specifically, they proposed to model the label probability of a bag
via the following mechanism:
\begin{equation}
p(y|X) = \rho \left( \mathbf{z} \right) = \rho \left( \sum_{k=1}^{K} \alpha_k \mathbf{h}_k \right) = \rho
\left( \sum_{k=1}^{K} \alpha_k \phi\left(\mathbf{x}_k\right) \right)
\label{eq:attentionmil}
\end{equation}
where $\phi:\mathbb{R}^{N} \mapsto \mathbb{R}^M$ and
$\rho:\mathbb{R}^M \mapsto [0, 1]$. The coefficients 
$\alpha_k$ in Eq. \ref{eq:attentionmil} can be computed 
via the additive attention \cite{bahdanau2014neural} mechanism of Eq. \ref{addatt},
which introduces the learnable parameters 
$\mathbf{V} \in \mathbb{R}^{L \times M}, \mathbf{w} \in \mathbb{R}^{L \times 1}$.
\begin{equation}
\alpha_k = \frac{\exp{(\mathbf{w}^T \tanh{(\mathbf{V}\mathbf{h}_k^T}) } )}
{\sum_{k=1}^{K}\exp{(\mathbf{w}^T \tanh{(\mathbf{V}\mathbf{h}_k^T}) })}
\label{addatt}
\end{equation}

In this work, we adopt the attention-based multiple-instance
scheme for the purposes of detecting rDR in a fundus image.
In a nutshell, we propose the following processing pipeline:
\begin{enumerate}
	\item Pre-process the input image $I$ according to Section \ref{sec:preproc} 
	\item Compute its bag of patches representation as described in Section \ref{sec:bop}.
	\item Transform the bag of patches $X=\{\mathbf{x}_1,
          \mathbf{x}_2, \dots, \mathbf{x}_K\}$ with 
          $\mathbf{x}_i \in \mathbb{R}^{d \times d \times 3}$ to a bag of features
          $H=\{\mathbf{h}_1, \mathbf{h}_2, \dots, \mathbf{h}_K\}$ 
          with $\mathbf{h}_i \in \mathbb{R}^M$, by
	applying the function $\phi$ to each patch.
	\item Apply attention pooling to arrive 
	at a global image representation 
	vector $\mathbf{z} = \sum_{k=1}^{K} \alpha_k \mathbf{h}_k$
	\item Estimate the rDR probability for $I$ 
	by computing $\rho (\mathbf{z})$ (where $\rho$ 
	is modelled by a fully-connected network).
\end{enumerate}

\begin{figure*}[!ht]
	\centering
	\subfloat[Proliferative DR\label{cut1}]{
		\includegraphics[scale=0.5,width=0.238\linewidth]{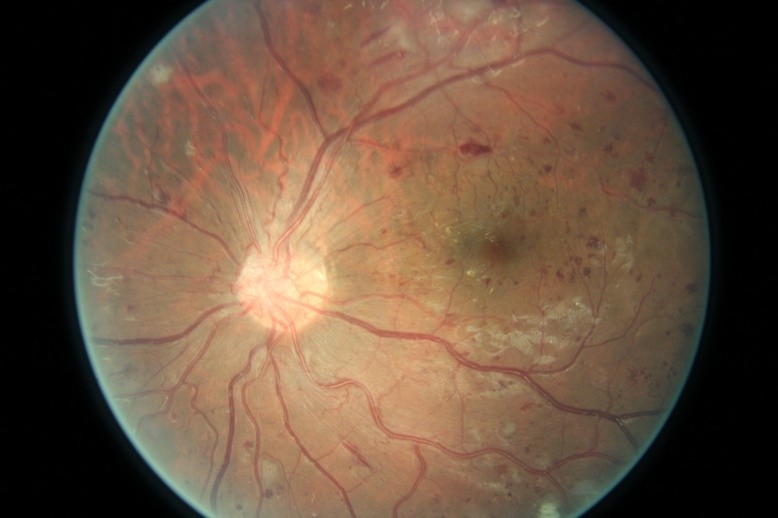}
	}
	\subfloat[No DR]{
		\includegraphics[scale=0.5,width=0.238\linewidth]{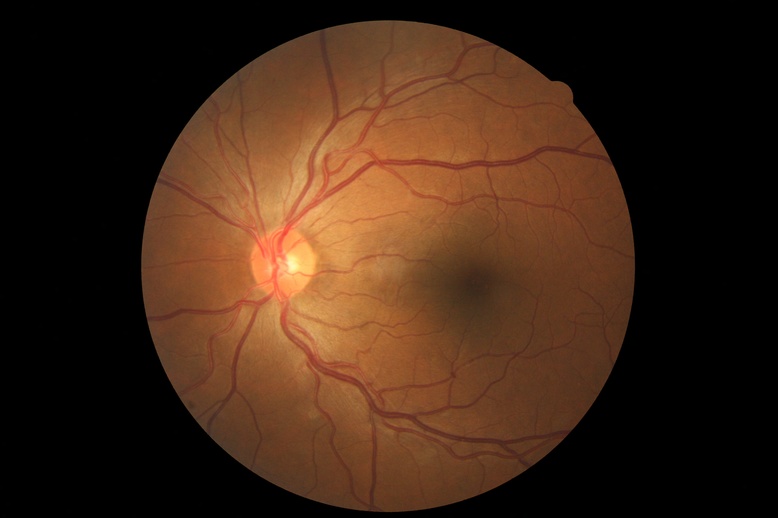}
	}
	\subfloat[Proliferative DR\label{cut2}]{
		\includegraphics[scale=0.5,width=0.238\linewidth]{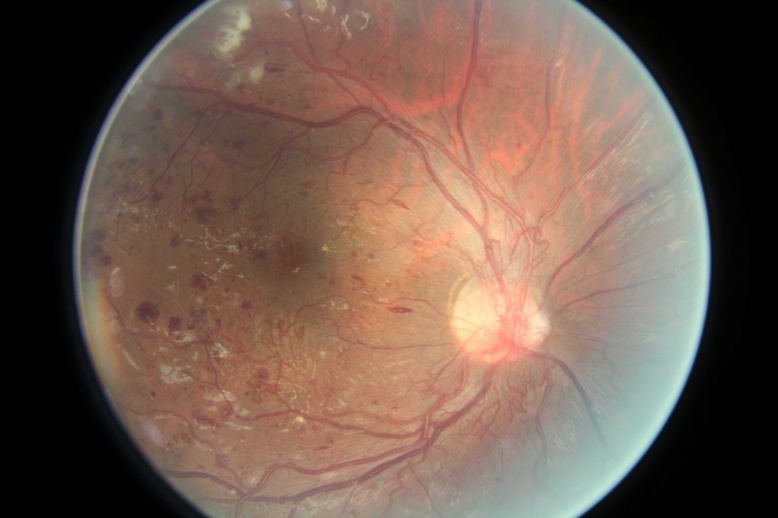}
	}
	\subfloat[Mild DR]{
		\includegraphics[scale=0.5,width=0.238\linewidth]{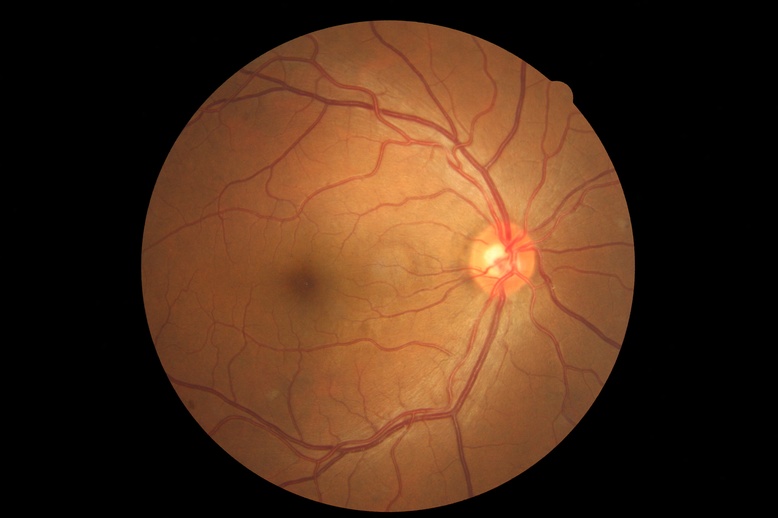}
	}
	
	\hspace{0mm}
	\subfloat[Proliferative DR]{
		\includegraphics[scale=0.5,width=0.238\linewidth]{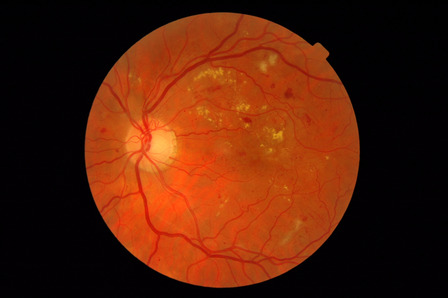}
	}
	\subfloat[Severe DR]{
		\includegraphics[scale=0.5,width=0.238\linewidth]{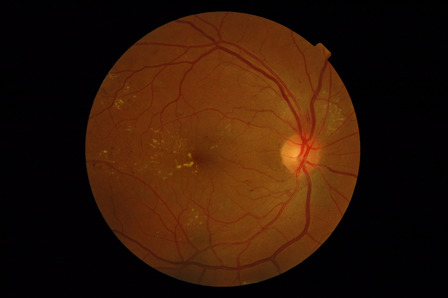}
	}
	\subfloat[No DR]{
		\includegraphics[scale=0.5,width=0.238\linewidth]{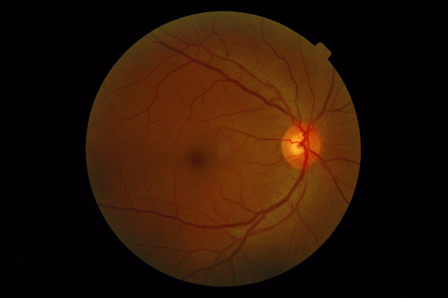}
	}
	\subfloat[Moderate DR]{
		\includegraphics[scale=0.5,width=0.238\linewidth]{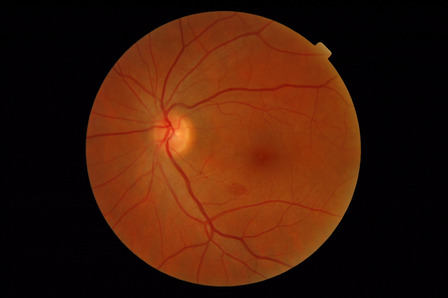}
	}
	
	
	\caption{Sample images from the Kaggle-EyePACS dataset (top row)
		and the Messidor-2 dataset (bottom row) with
		their corresponding DR grades.
		The first row also showcases
		the two types of fundus images contained in Kaggle-EyePACS,
		i.e. images where the circular disk of the eye
		is fully captured (right column) or images where the circular disk
		is partially clipped (left column).
	}\label{data_samples}
\end{figure*}

One attractive property of the MIL attention approach is that
all transformations can be modelled using deep neural networks
and the resulting model can still be trained in an end-to-end manner.
This would not be possible if some other commonly used pooling operator, such as
Bag of Features or Fisher Vectors, was used.
For our specific problem, we elect to use a high 
capacity CNN (ResNet-18) \cite{he2016deep} to model 
the feature extraction function $\phi$ and a single 
fully-connected layer to model the final classifier $\rho$.
The attention mechanism is implemented via 
a fully-connected network with 2 layers 
that correspond to the parameters $\mathbf{V}$ and $\mathbf{w}$ of Eq. \ref{addatt}.
A schematic overview of the proposed system is given in Fig. \ref{pipeline}.

\subsection{Interpretable attention heatmaps}\label{sec:heatmaps}
After training, we can use the attention mechanism to visualize
the fundus regions that affect the model's decision. 
As we saw in section \ref{sec:mil}, in order to classify an image,
we assign a weight to each patch
using the attention mechanism. These weights offer an implicit
way of identifying which image regions contribute the 
most to the model's prediction.
Ideally, in true positive cases, the model should
assign high weights $\alpha_k$ to patches that contain DR
lesions, such as haemorrhages and exudates, while in true 
negative cases, all patches should receive similar
weights. Such a heatmap could also assist in understanding the reasons
behind misclassifications, either false positives or false negatives
and enable targetted interventions in the learning pipeline.

To produce the attention heatmap, we start with an initially 
zeroed single-channel image of same height and width as the pre-processed images. 
This auxiliary image will be used to accumulate the attention values:
we iterate the patches and add their assigned weights to the
pixels of the accumulator image that correspond to the location of each
patch in the original image. The granularity of the pixel-level assignment can
be controlled via the patch overlap parameter $t$: larger overlap between the extracted
patches leads to more granular and aesthetically pleasing visualizations, while
smaller overlap speeds up the computations but results in coarser visualizations.
The values of the auxiliary image are then linearly mapped to the $[0, 1]$ range
to produce a proper heatmap.
Examples of such heatmaps are given in Fig. \ref{fig:visualizations}.

\section{Datasets}\label{sec:datasets}
\subsection{Kaggle-EyePACS}
The Kaggle Diabetic Retinopathy Detection \cite{dataset} challenge dataset
consists of high-resolution retina images captured under a 
variety of imaging conditions. It contains $88.702$ RGB images of differing
resolutions (from $433\times289$ up to $5184\times3456$) that are partitioned
in a training set of $35.126$ and a test set of $53.576$ images. A clinician
has graded all images according to the ICDR scale. The DR 
grade distribution can be seen in Table \ref{tab:grade_dists}.
Some indicative images from this dataset can be seen in Fig. \ref{data_samples}.

It is estimated \cite{rakhlin2018diabetic} that 
$25\%$ of the Kaggle-EyePACS images are ungradable because
they contain artifacts (loss of focus, under or overexposure
etc.) that prevent a medical expert from assessing their DR grade. 
In fact, it is quite common for DR image graders to first evaluate the 
image quality and proceed with the actual DR grading only if they
find it acceptable.
The authors of \cite{voets2018replication} suggested
that image quality estimation does not require medical expertise
and can, therefore, be carried out by non-experts 
They went on to manually evaluate 
the quality of all images in the Kaggle dataset, based on instructions provided 
to the professional DR graders for performing the same task 
in \cite{gulshan2016development}. In doing so, they rejected about $19.9\%$ of
the images, resulting in a filtered dataset of $71.056$ images. 
In our experiments, we adopt these gradability
estimates and consider only the images that were deemed of adequate quality.

\subsection{Messidor-2}
The Messidor-2 dataset \cite{abramoff2013automated} 
contains $1748$ fundus images captured with a Topcon TRC
NW6 camera in three different resolutions:
$1440\times960$, $2240\times1488$ and $2304\times1536$.
Grades for DR and image quality are provided
by a panel of 3 fellowship-trained retina specialists \cite{krause2018grader}.
Images that caused disagreements among specialists were re-examined and
adjudication sessions were carried out until consensus was reached.
As a result, in terms of label quality, Messidor-2 can be considered less
noisy than the Kaggle-EyePACS dataset, in which DR grades 
are provided by a single DR grader. 
The DR grade distribution for this dataset is given in Table \ref{tab:grade_dists},
while sample images are provided in Fig. \ref{data_samples}. 

\begin{table*}[!ht]
	\centering
	\caption{Distribution of the 5 DR grades in the images of the 
		Kaggle-EyePACS and Messidor-2 datasets. Kaggle-EyePACS images
              have been graded by a single expert, while Messidor-2 images
            by a panel of 3 experts via an adjudication process.}
	\begin{tabular*}{\textwidth}{@{\extracolsep{\fill}} l|cc|ccc|ccc}
		\toprule
		
		\multicolumn{1}{c}{} & \multicolumn{2}{c}{No rDR} & \multicolumn{3}{c}{rDR} & \multicolumn{3}{c}{Train-Test Split} \\
		\midrule
		\textbf{Dataset} & \textbf{No DR} & \textbf{Mild} & \textbf{Moderate} & \textbf{Severe} & \textbf{Proliferative} & Train & Test & Total \\
		\midrule
		Kaggle-EyePACS (all) & 65.343 (73.6\%) & 6.205 (6.9\%) & 13.153 (14.8\%) & 2.087 (2.3\%) & 1.914 (2.1\%) & 35.126 & 53.576 & 88.702 \\
		Kaggle-EyePACS (gradable) & 52.649 (74.1\%) & 5.073 (7.1\%) & 10.279 (14.4\%)& 1.596 (2.2\%) & 1.372 (1.9\%) & 28.098 & 42.871 & 70.969 \\
		\midrule
		Messidor-2 (gradable) & 1.017 (58.3\%) & 270 (15.4\%) & 347 (19.8\%) & 75 (4.3\%)& 35 (2.0\%) & & 1.744 & 1.744\\
		\bottomrule
	\end{tabular*}
	\label{tab:grade_dists}
\end{table*}

\subsection{IDRiD}
The Indian Diabetic Retinopathy Dataset (IDRiD) \cite{h25w98-18} contains fundus images
captured during real clinical examinations in an eye clinic in India
using a Kowa VX fundus camera. The captured images have a
50\textdegree  field of view with a resolution of $4288\times2848$.
The images are separated into 3 parts,
corresponding to 3 different learning tasks and
accompanied by the respective types of ground-truth.
The first part is designed for the development
of segmentation algorithms and contains 81 images (54 train set -
27 test set) with pixel-level annotations of
DR lesions (microaneurysms, haemorrhages, hard and soft exudates)
and the optical disk. The second part corresponds to a DR grading task and contains 516 images
divided into train set (413 images) and test set (103 images) with DR and Diabetic Macular
Edema (DME) severity grades.
Finally, the third part corresponds to a localization task and contains 516
images with the pixel coordinates of the optic disk center and fovea center (again
split in a 413 train and 103 test set).

\section{Experimental Evaluation}\label{sec:experiments}

\subsection{Implementation details}
We perform extensive experiments to assess the proposed model's ability 
to detect rDR in fundus images.
For training, we use the gradable (according to \cite{voets2018replication}) 
images in the Kaggle-EyePACS training split, while for evaluation
we use the Kaggle test split and Messidor-2 datasets.
Before any training, we randomly sample $3000$ images from the Kaggle training
and set them aside for use as a validation set.
For training, we use the standard binary cross-entropy loss.
Assuming that $\hat{p}\textsubscript{data}$ 
denotes the empirical data-label distributions defined by the training set,
the cross-entropy loss is defined as:
\begin{equation} \label{eq:loss_function}
\mathcal{L} = \mathop{-\mathbb{E}}_{X,y\textsubscript{rdr} 
    \sim \hat{p}\textsubscript{data}} 
    \big[y\textsubscript{rdr}  \log(p\textsubscript{model}(y\textsubscript{rdr} | X)\big]
\end{equation}

\begin{figure}[!h]
  \centering
\begin{tikzpicture}
  \pgfplotsset{every tick label/.append style={font=\small}}
\begin{axis}[
    xlabel={Random patches per bag / Patch overlap},
    ylabel={Validation set AUC},
    xmin=0, xmax=120,
    ymin=0.9, ymax=1,
    xtick={5,10,20,50,100},
    xticklabels={5,10,20/0,50/0.25,100/0.5},
    legend pos=south east,
    ymajorgrids=true,
    xmajorgrids=true,
    grid style=dashed,
    cycle list name=exotic,
]

\addplot+[
    ]
    coordinates {
      (5,0.913)(10,0.947)(20,0.954)(50,0.961)(100, 0.958)
    };
    \addlegendentry{Random patch selection}

\addplot+[
    ]
    coordinates {
      (20,0.957)(50,0.953)(100,0.955)
    };
    \addlegendentry{Deterministic patch selection}

\end{axis}
\end{tikzpicture}
\caption{AUC scores on the Kaggle validation set 
  for random patch selection 
  and deterministic patch selection.
  The alternative labels of the x-axis, when given, correspond to 
  the patch overlap value for the deterministic patch selection policy.
  }
\label{fig:patch_policy}
\end{figure}
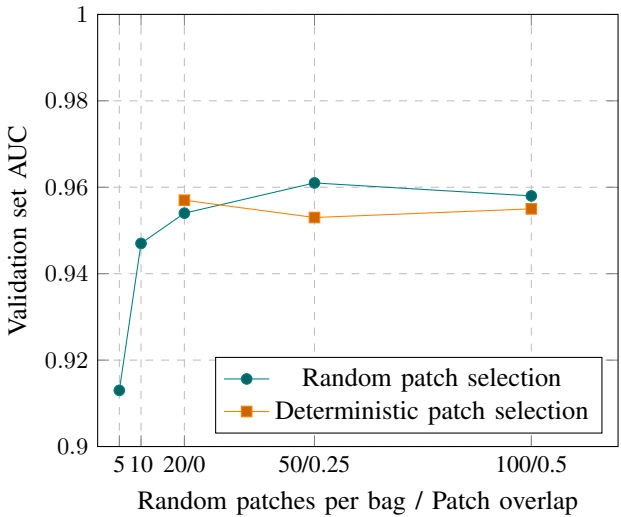

The work of \cite{raghu2019transfusion} showed that transferring weights from
the natural to the medical image domain is not necessary
to achieve high classification performance. However, it does speed up convergence
and thus, we initialized the function $\phi$ (ResNet-18) with pre-trained 
ImageNet weights.
During training we perform image augmentation by randomly 
shifting, flipping, scaling and rotating
the image and by randomly applying small perturbations to its brightness,
contrast, hue and saturation.
We convert the image to its bag of patches representation
by randomly selecting $K=50$ random patches from the pool of all
possible image patches. 
This value was selected based on its superior classification
performance (Fig. \ref{fig:patch_policy}) on the validation 
set.
At test time, instead, we extract patches on a regular grid
with a patch overlap rate of $0.75$.
The function $\phi$ transforms an input patch to a feature vector of
length $M=128$, while an attention module of dimension $L=32$ is used to 
produce the pooled image representation $z$.
We use the Adam optimizer with a base learning rate of $3\cdot 10^{-4}$ for $60$ epochs
and the suggested $b_1, b_2$ parameters. After training, we keep the model 
instance that achieved the highest performance on the validation set.

\subsection{Classification experiments}

\begin{table*}[!ht]
	\centering
	\caption{Comparison of classification performance between the 
        proposed approach and the most notable works in the related literature for
        the Kaggle-EyePACS and Messidor-2 datasets. For Kaggle we report 
        separate methods that use the official training/test split (first part of
        the table) from works that use custom data splits, typically in the range of
        $80\%$ training - $20\%$ testing (second part of the table). Sensitivity/specificity
        values for the operating points of high sensitivity/specificity are reported when
        available.}
	\begin{tabular}{cclc|cc|cc}
		\toprule
		\multicolumn{1}{c}{} & \multicolumn{1}{c}{} & \multicolumn{1}{c}{} & \multicolumn{1}{c}{} & \multicolumn{2}{c}{High sensitivity point} & \multicolumn{2}{c}{High specificity point} \\
		\toprule
		\textbf{Trained on} & \textbf{Evaluated on} & \textbf{Method} & \textbf{AUC} & \textbf{Sensitivity} & \textbf{Specificity} & \textbf{Sensitivity} & \textbf{Specificity} \\
		\midrule
		\multirow{5}{*}{\parbox{2cm}{\centering Kaggle train}} & 
		\multirow{5}{*}{\parbox{2cm}{\centering Kaggle test}}
		& Deep MIL Attention & \textbf{0.957} & 0.951 & 0.753 & 0.843 & 0.951 \\
		& & Leibig et al. \cite{leibig2017leveraging} & 0.927 & 
		\multicolumn{2}{c}{NA} & \multicolumn{2}{c}{NA} \\
                & & Rakhlin et al. \cite{Rakhlin225508} & 0.920 & 0.920 & 0.720 & 0.800 & 0.920 \\
		& & Pires et al. \cite{pires2019data} & 0.946 & 
		\multicolumn{2}{c}{NA} & \multicolumn{2}{c}{NA} \\
		& & O\_o \cite{graham2015kaggle} & 0.951 & 
		\multicolumn{2}{c}{NA} & \multicolumn{2}{c}{NA} \\

		\midrule[.01em]
		\multirow{3}{*}{\parbox{2.5cm}{\centering Kaggle custom train}} & 
		\multirow{3}{*}{\parbox{2.5cm}{\centering Kaggle custom test}}
		& Voets et al. \cite{voets2018replication} & 0.941 & 0.899 & 0.838 & 0.834 & 0.901 \\
		& & Zhou et al. \cite{zhou2017deep} & 0.925 
		& \multicolumn{2}{c}{NA} & \multicolumn{2}{c}{NA} \\
		& & Quellec et al. \cite{quellec2017deep} & 0.954 & 
		\multicolumn{2}{c}{NA} & \multicolumn{2}{c}{NA} \\
		
		\midrule
		Kaggle train & 
		\multirow{7}{*}{\parbox{2cm}{\centering Messidor-2}} 
		& Deep MIL Attention & 0.976 & 0.954 & 0.845 & 0.856 & 0.953 \\
		Kaggle custom train & & Voets et al. \cite{voets2018replication} & 0.800 & 0.737 & 0.697 & 0.697 & 0.764 \\
		Kaggle custom train & & Zhou et al. \cite{zhou2017deep} & 0.960 & 
		\multicolumn{2}{c}{NA} & \multicolumn{2}{c}{NA} \\
		Augmented EyePACS & & Gulshan et al. \cite{gulshan2016development} & 
		\textbf{0.990} & 0.961 & 0.939 & 0.870 & 0.985 \\
		EyePACS & & Gargeya et al. \cite{gargeya2017automated} & 0.940 & 
		\multicolumn{2}{c}{NA} & \multicolumn{2}{c}{NA} \\
		Private & & Abramoff et al. \cite{abramoff2013automated}* & 0.935 & 
		0.968 & 0.594 & \multicolumn{2}{c}{NA} \\
		Private & & Abramoff et al. \cite{abramoff2016improved}* & 0.980 & 
		0.968 & 0.870 & \multicolumn{2}{c}{NA} \\
		Kaggle train & & Rakhlin et al. \cite{Rakhlin225508} & 0.970 & 
                0.990 & 0.710 & 0.870 & 0.920 \\
		\bottomrule
	\end{tabular}
	\label{results}
\end{table*}
For measuring classification performance we use the \emph{Area Under the Receiver Operating Curve (ROC AUC)},
a metric commonly used in the related literature.
We also compute the model's sensitivity and specificity metrics 
at the operating points of high specificity ($>0.9$) and high sensitivity ($>0.9$).
We report the performance metrics of a 5 model ensemble 
on both the Kaggle-EyePACS test set and Messidor-2
in Table \ref{results}, along with a comparison to related works. 
Ideally, such a comparison would include methods
that were trained and evaluated on the same datasets and under the same conditions.
However, there is a widely inconsistent use of datasets and evaluation metrics throughout 
the rDR classification literature \cite{stolte2020survey}, with different methods using different
datasets for training/testing or even custom training/test splits. 
For instance, \cite{voets2018replication} and \cite{zhou2017deep} use Kaggle splits that favor
significantly larger training sets, as opposed to
the official splits in which the test set is $\sim$1.5 times larger than the training set.
After taking this into consideration, we elect to compare our method to the most prominent works of the rDR literature
that use the same publicly-available datasets. More specifically, in the first part
of Table \ref{results} we include methods that were trained and evaluated on Kaggle-EyePACS,
using either official or custom splits. Then, in the second part we include
methods that were trained with any dataset and evaluated on Messidor-2.

As we can see, in terms of rDR classification performance, our method perform on par with 
the state-of-the-art literature. In fact, when evaluated on the Kaggle test set, our approach
outperforms the alternatives in terms of AUC score, even those that use larger training sets.
In Messidor-2, our method is only outperformed by \cite{gulshan2016development}. However,
their model was trained on a much larger and better annotated dataset ($>$100k images
annotated by a panel of specialists, in contrast to our $\sim$28k images annotated by
a single expert). \cite{abramoff2016improved} also reports slightly better AUC on Messidor-2,
but their score corresponds to a slightly different task that is to predict rDR given both
eye images for a subject, in contrast to our model that operates on single images.

\begin{figure*}[!ht]
  \centering

  \subfloat[rDR positive - rDR probability $0.99$]{
    \includegraphics[scale=0.5,width=.5\linewidth]{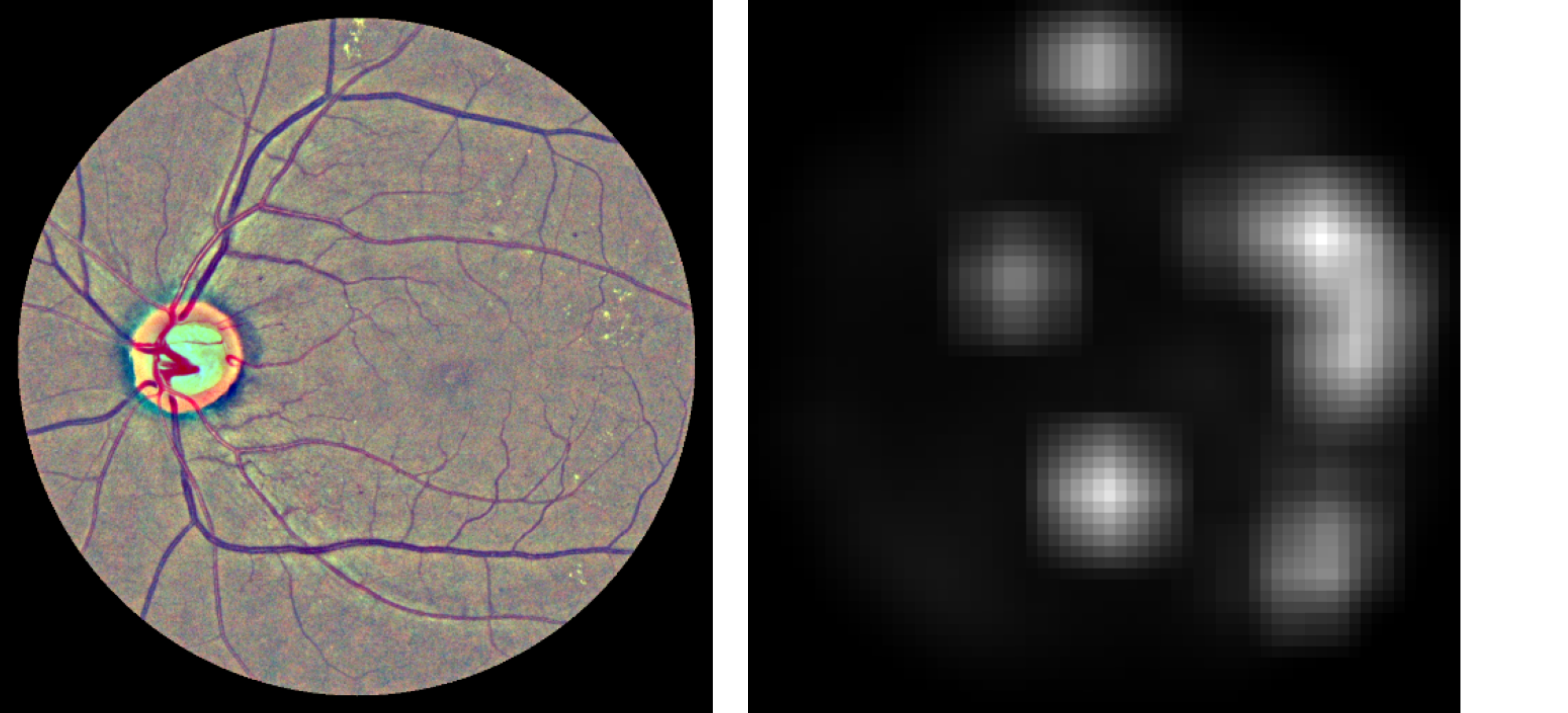}
  }
  \subfloat[rDR positive - rDR probability $0.99$]{
    \includegraphics[scale=0.5,width=.5\linewidth]{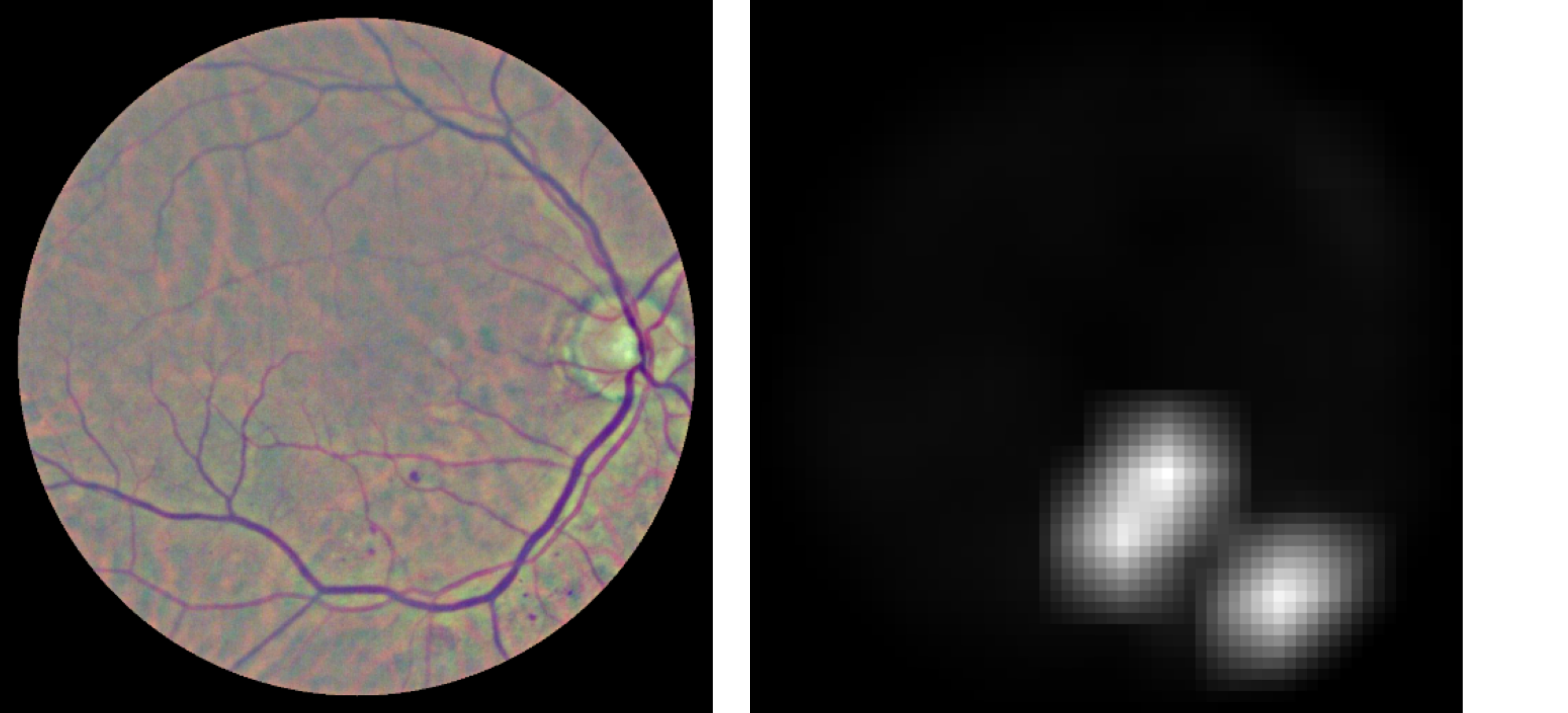}
  }
  \hspace{0mm}
  \subfloat[rDR positive - rDR probability $0.99$]{
    \includegraphics[scale=0.5,width=.5\linewidth]{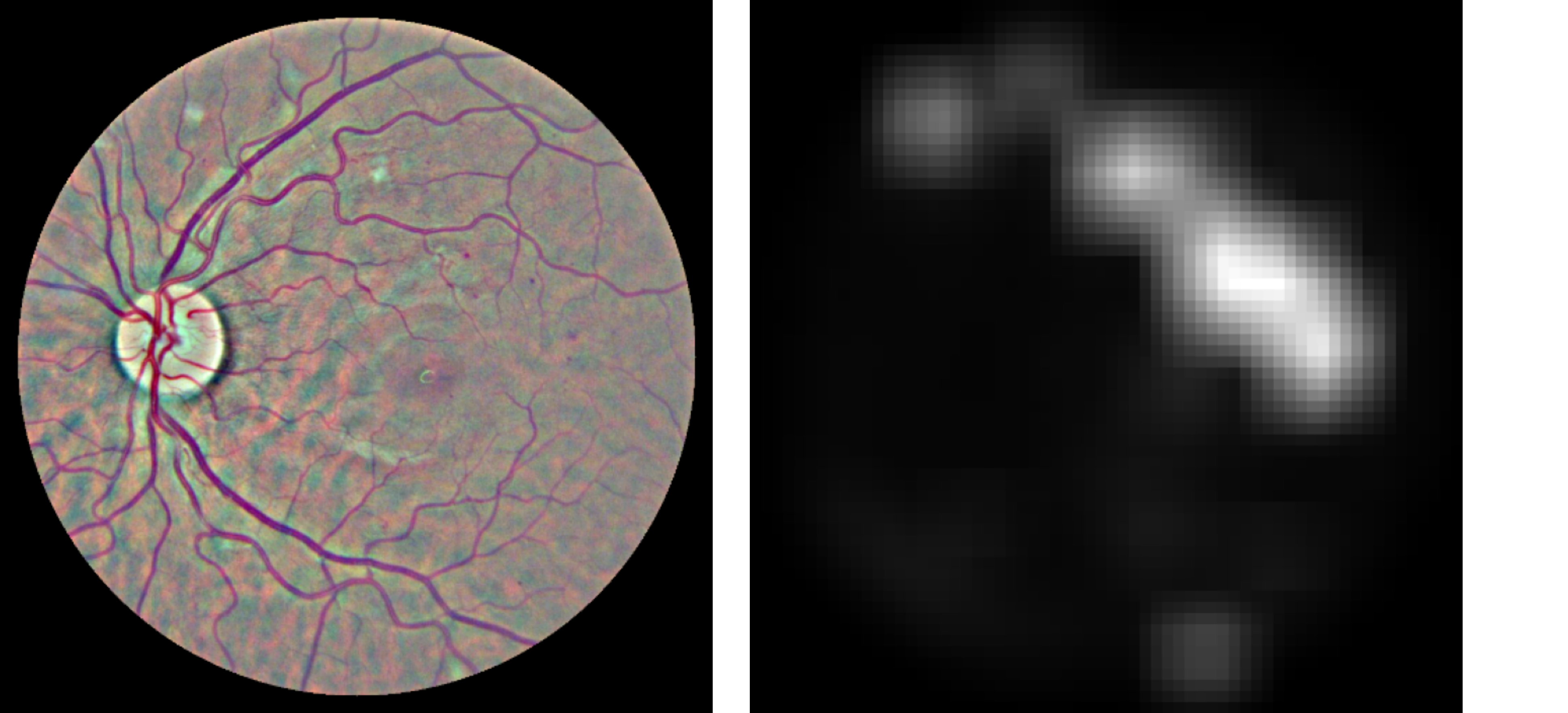}
  }
  \subfloat[rDR negative - rDR probability $0.97$\label{fig:fp1}]{
    \includegraphics[scale=0.5,width=.5\linewidth]{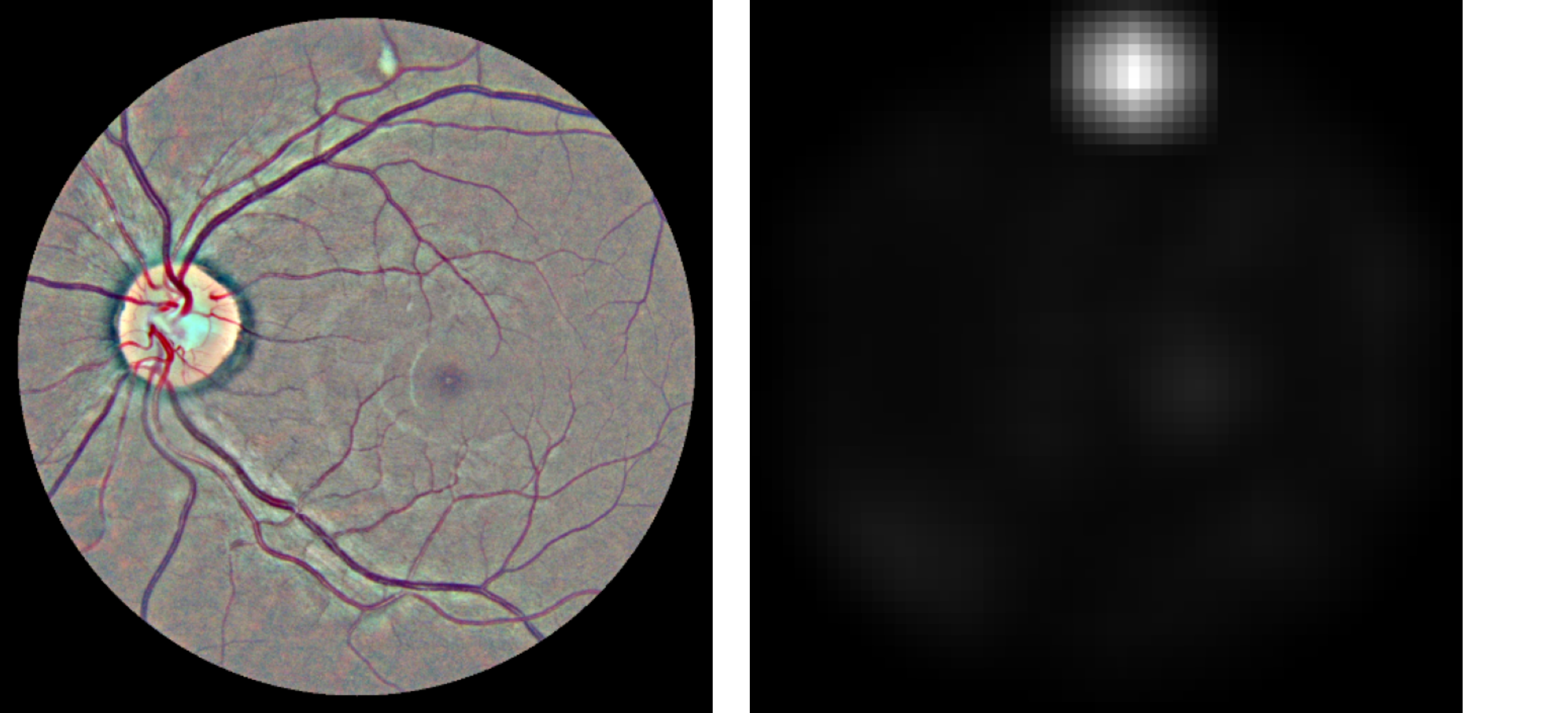}
  }
  \hspace{0mm}
  \subfloat[rDR negative - rDR probability $0.96$\label{fig:fp2}]{
    \includegraphics[scale=0.5,width=.5\linewidth]{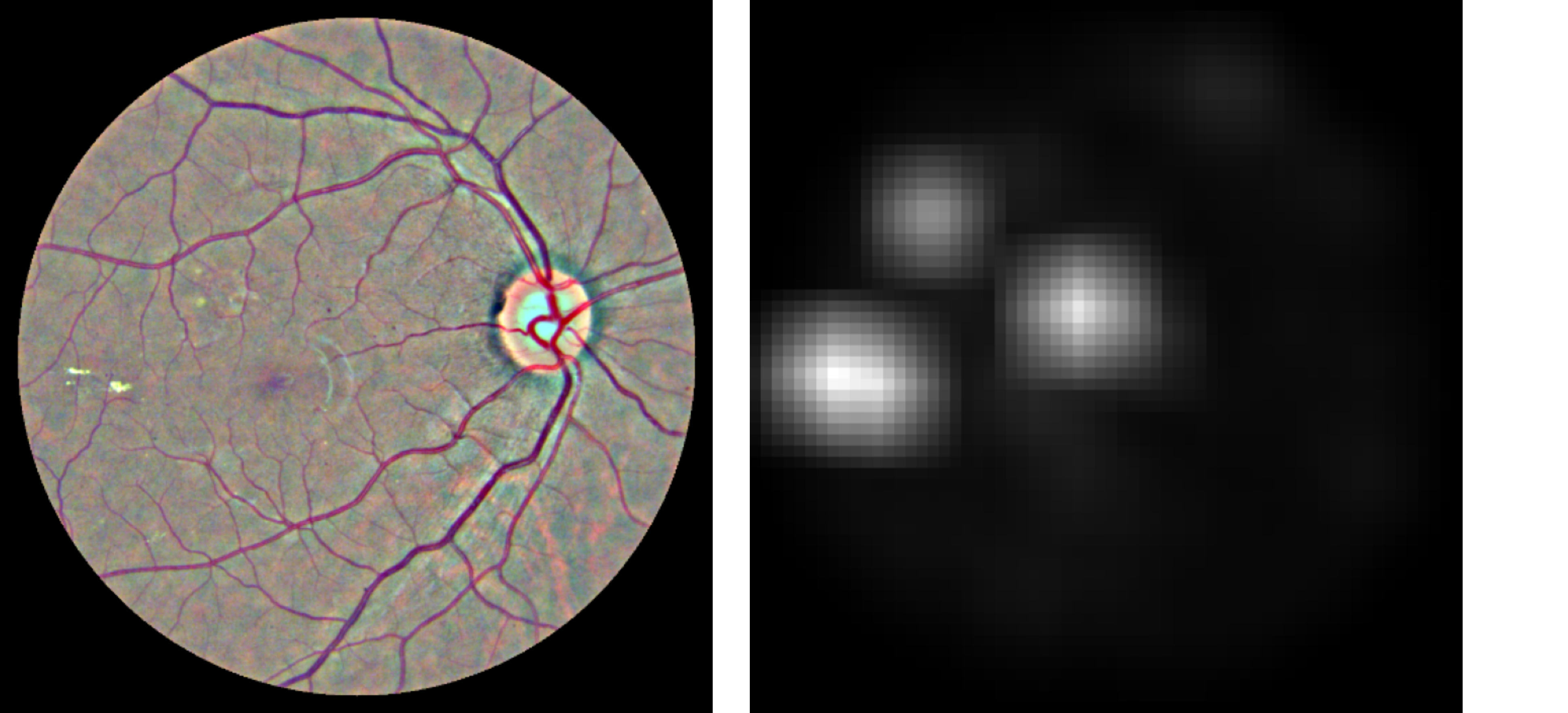}
  }
  \subfloat[rDR negative - rDR probability $0.82$\label{fig:fp3}]{
    \includegraphics[scale=0.5,width=.5\linewidth]{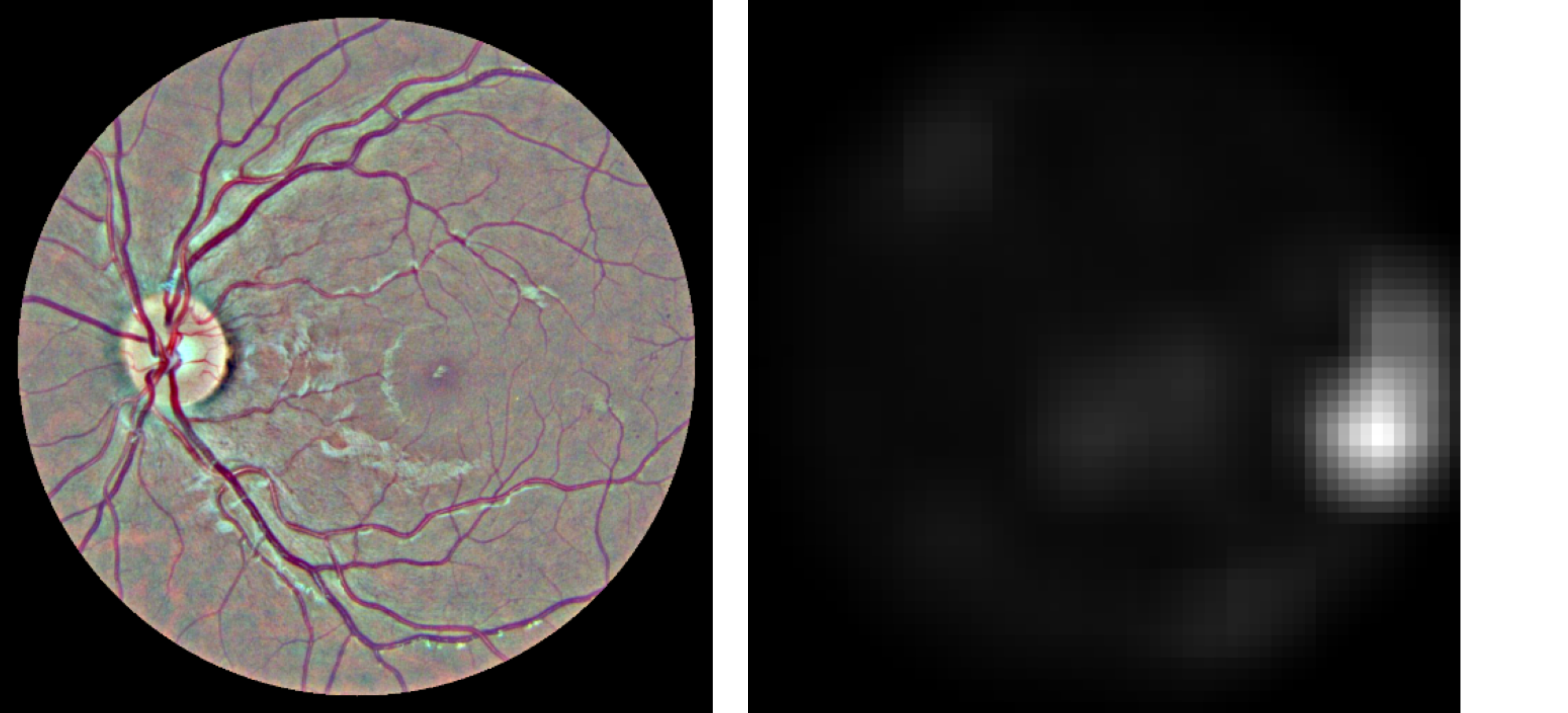}
  }
  
  \caption{Examples of attention heatmaps produced by the model
    for 6 Messidor-2 images that corresponded to both true-positive and 
  false-positive predictions.} \label{fig:visualizations}
\end{figure*}

\subsection{Attention heatmap evaluation}
Following the procedure of Sec. \ref{sec:heatmaps} we can
construct attention heatmaps that pinpoint the fundus areas that
the model focused on for arriving at a prediction.
Examples of such heatmaps for both correct and incorrect predictions in
the Messidor-2 dataset are presented in Fig. \ref{fig:visualizations}.
As we can see by inspecting the images in Fig. \ref{fig:visualizations},
the model seems to attend more to patches that contain artifacts 
resembling DR lesions, such as microaneurysms, haemorrhages and 
soft or hard exudates. Rather than relying on visual inspection
of the produced heatmap, we would like to evaluate its validity 
more quantitatively. To that end, we employ the first
part of the IDRiD dataset that contains 80 images with detailed,
pixel-level lesion annotations. 
Based on these images, we can verify the correlation
between per-patch attention weight and lesion existence
that Fig. \ref{fig:visualizations} suggests. To do so, 
we conduct two experiments. In the first experiment,
we use the attention weight assigned to each patch as 
a predictor of whether the patch contains any DR lesion. 
We compute the AUC and the Area Under Precision Recall Curve (AUPRC)
of the per-patch attention weight 
against a binary label that denotes lesion existence.
For the purposes of this experiment, we extract
patches on a regular grid with high patch overlap ($87.5\%$),
in order to enrich the pool of patches.
Any patch that contains even
a tiny amount of lesion (at least 1 pixel according to the
available ground-truth) is assigned to the positive class and otherwise
to the negative class. 
We report the aforementioned classification 
metrics against different labels, corresponding to 
microaneurysms, haemorrhages, exudates 
(we have merged soft and hard exudates in a single class, since
only a fraction of IDRiD images contain soft exudate ground-truth),
as well as all lesions combined, in Table \ref{tab:heatmap_clf}.
As we can see, the attention weight achieves good performance,
especially when the target label is produced by considering all
lesions together. This is to be expected, as the training procedure
allows the model to focus on what it considers relevant
to the task at hand and, as a result, it does not explicitly learn
to prioritize a specific lesion type over some other.

\begin{table}[!ht]
	\centering
	\caption{Resulting AUC and AUPRC scores when using the attention weight of a patch
        as a binary predictor of lesion presence. We report
        scores against different lesion types individually, as well
        as all lesions combined. Results are produced using images of the IDRiD dataset
        for which pixel-level lesion annotations are available.}
	\begin{tabular}{l|cccc}
		\toprule
                \multirow{2}{*}{\textbf{Lesion}} & \multirow{2}{*}{\parbox{1cm}{\centering Negative \\ patches}} & 
                \multirow{2}{*}{\parbox{1cm}{\centering Positive \\ patches}} & 
                \multirow{2}{*}{\textbf{AUC}} & 
                \multirow{2}{*}{\textbf{AUPRC}}\\
                                                 & & & & \\
		\midrule
                Microaneurysms & 163.735 & 96.185 & 0.760 & 0.611 \\
                Haemorrhages & 180.655 & 79265 & 0.745 & 0.528 \\
                Exudates & 162.637 & 97.283 & 0.749 & 0.591  \\
		\midrule
                Any lesion & 103.362 & 156.558 & 0.800 & 0.869 \\
		\bottomrule
	\end{tabular}
	\label{tab:heatmap_clf}
\end{table}

\begin{figure*}[!ht]
  \centering

  \subfloat[]{
    \includegraphics[width=.33\linewidth]{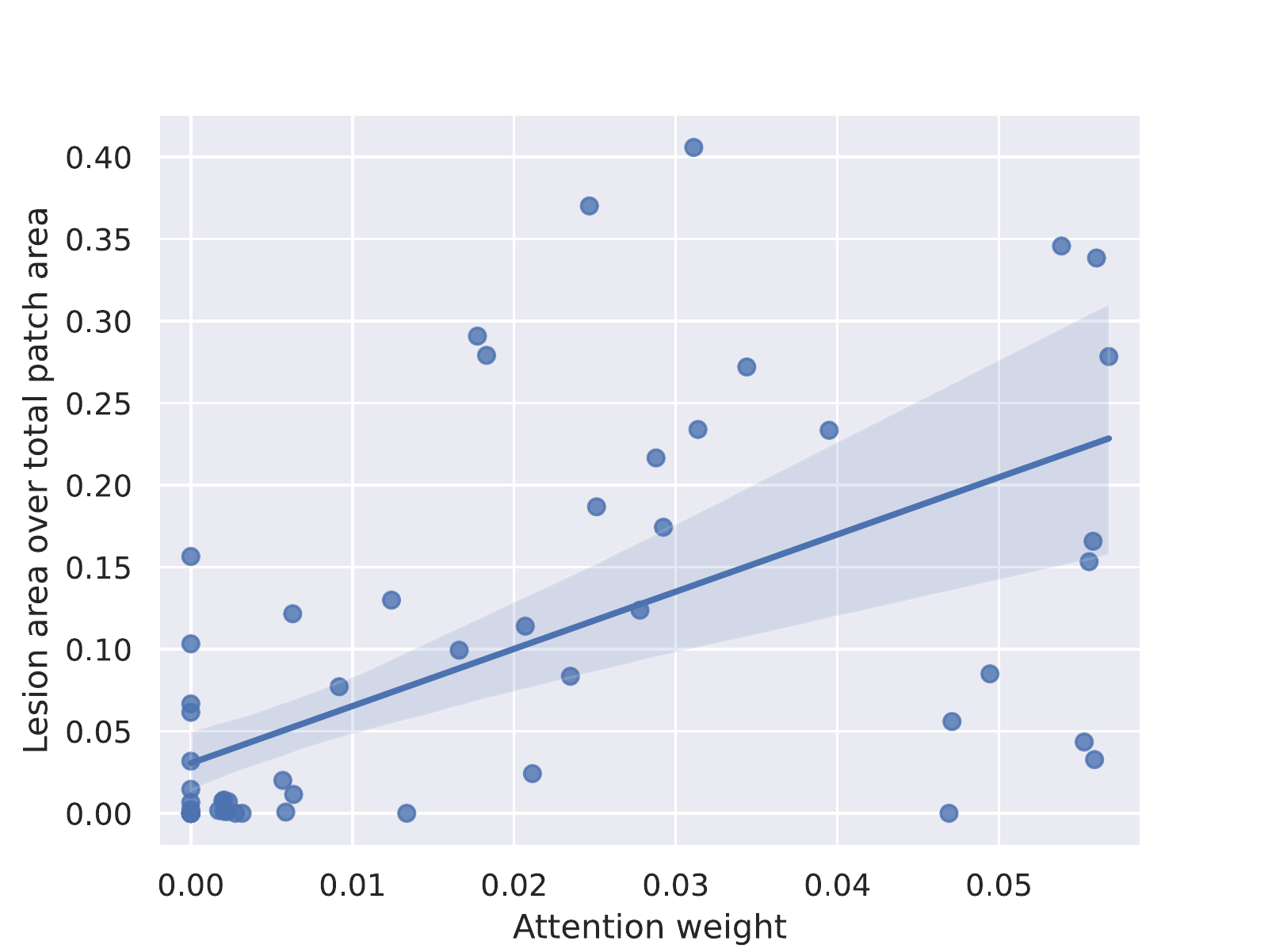}
  }
  \subfloat[]{
    \includegraphics[width=.33\linewidth]{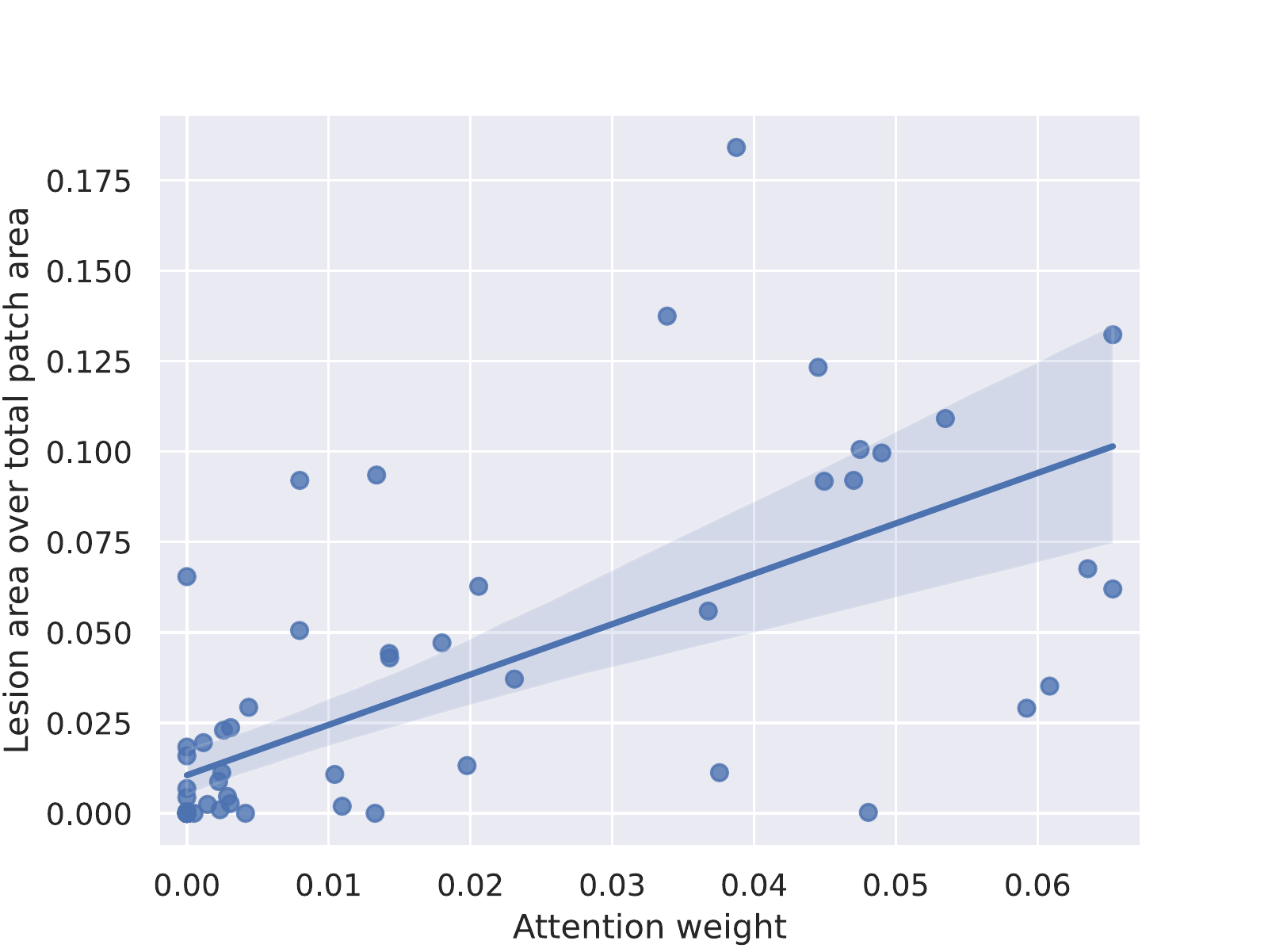}
  }
  \subfloat[]{
    \includegraphics[width=.33\linewidth]{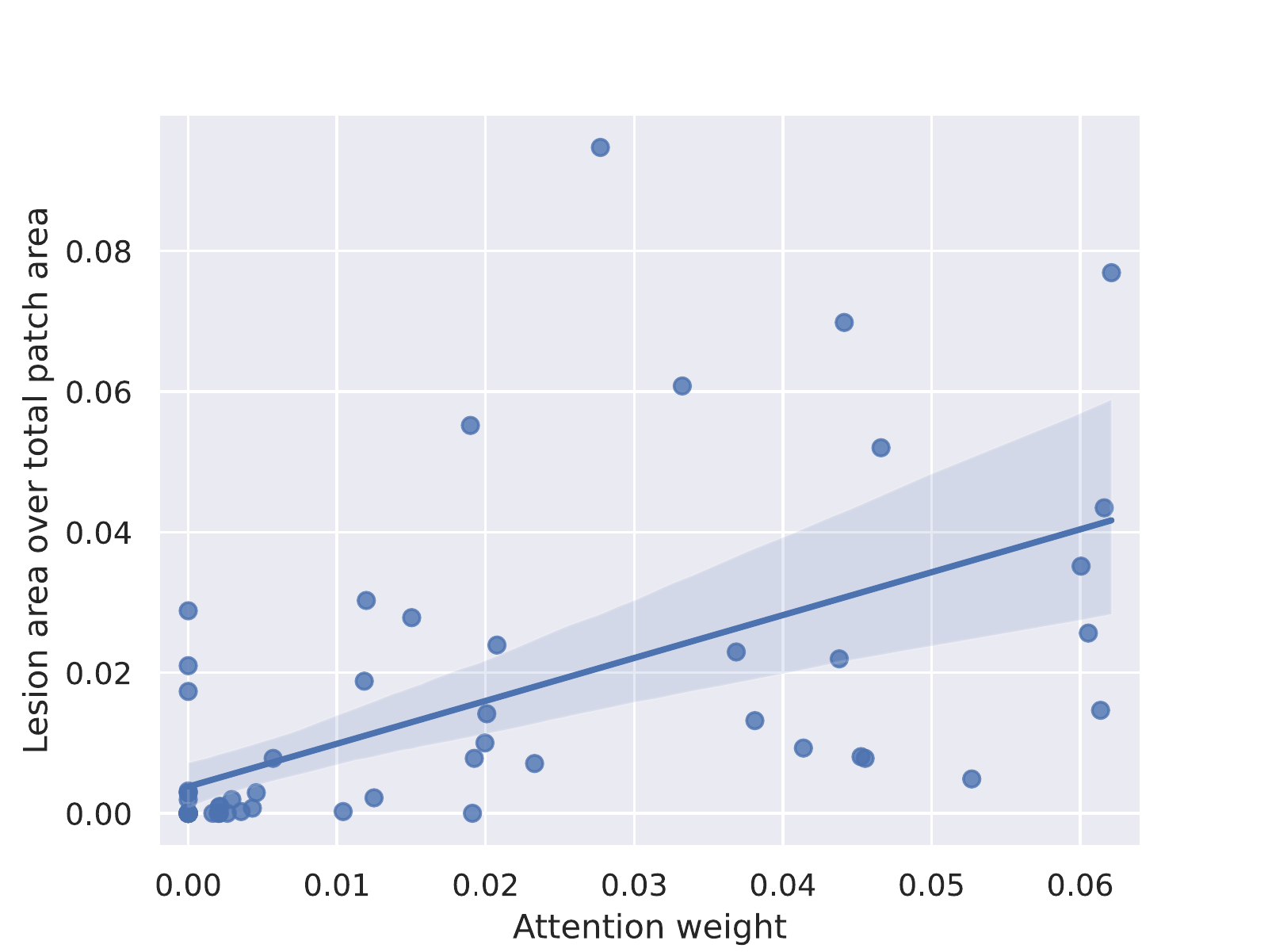}
  }
  \hspace{0mm}
  \subfloat[]{
    \includegraphics[width=.33\linewidth]{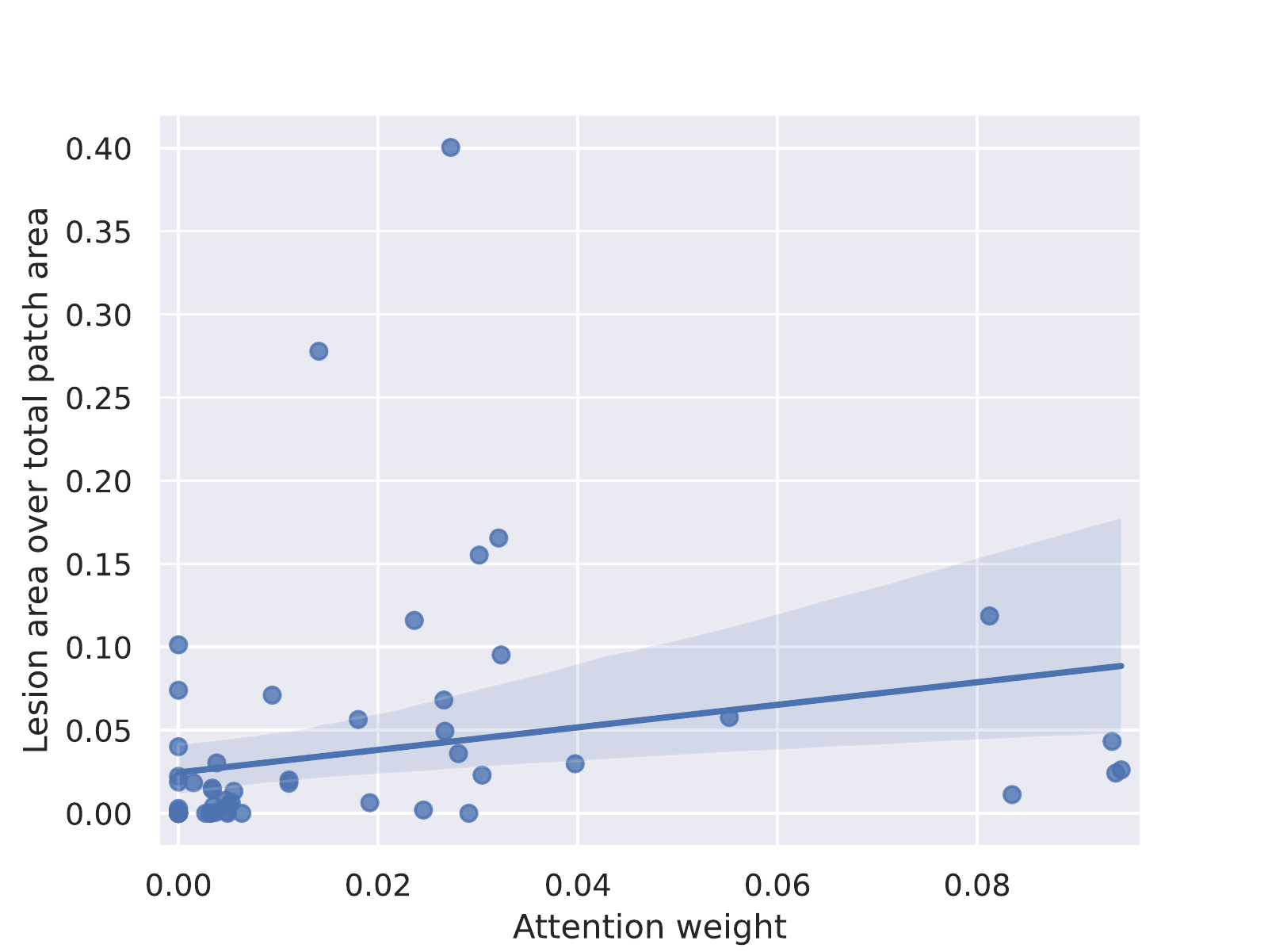}
  }
  \subfloat[]{
    \includegraphics[width=.33\linewidth]{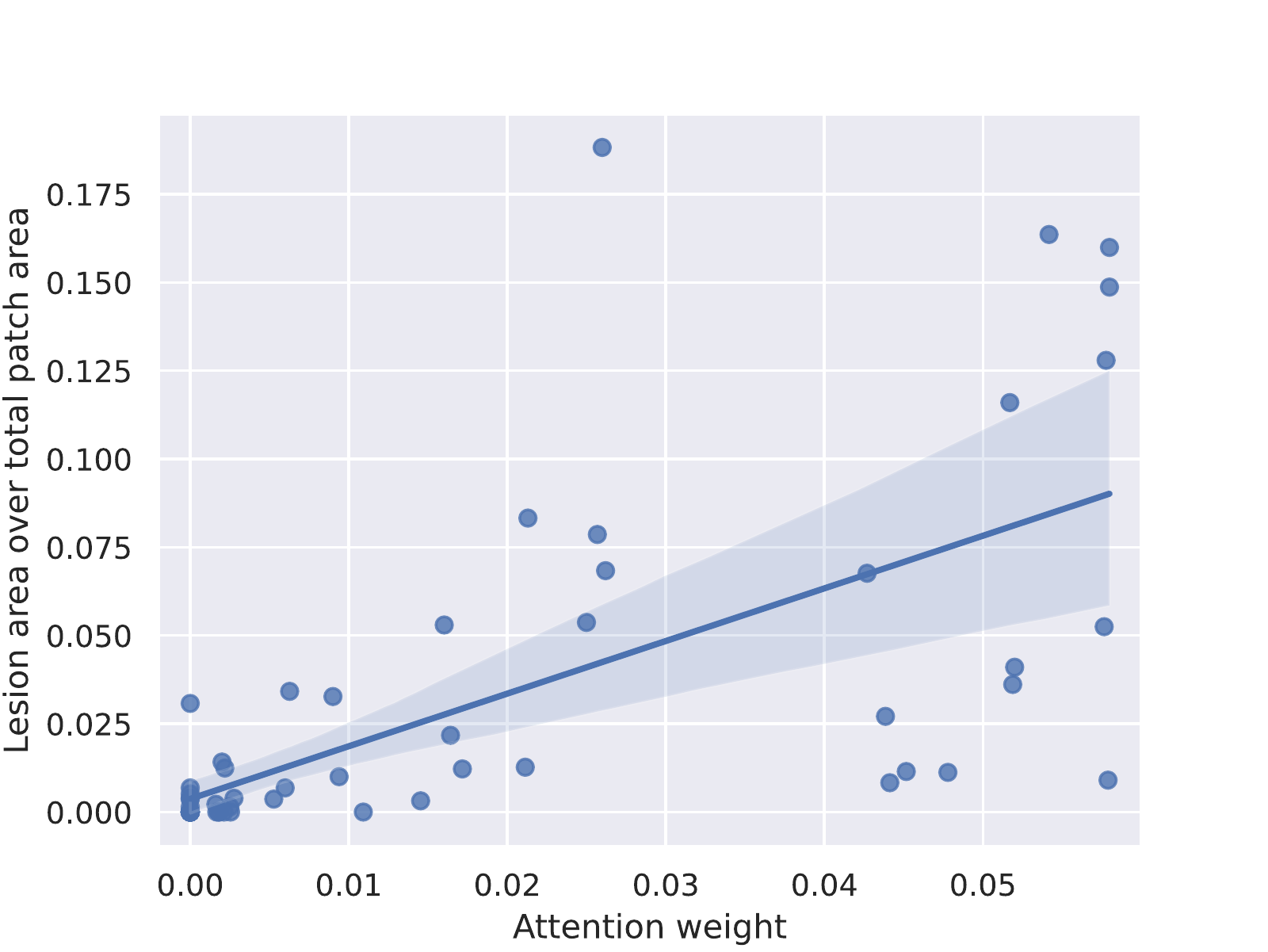}
  }
  \subfloat[]{
    \includegraphics[width=.33\linewidth]{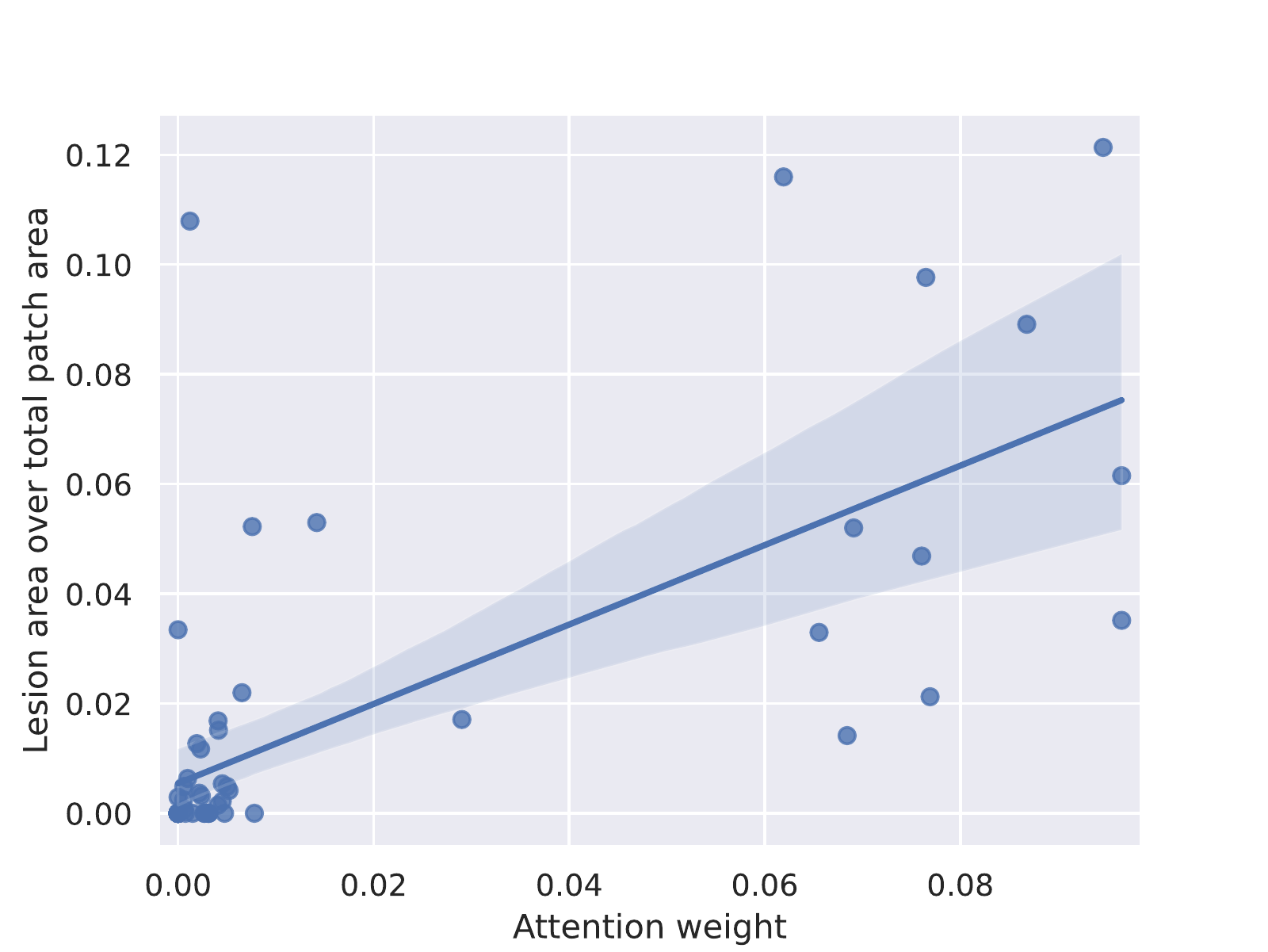}
  }
  \hspace{0mm}
  \subfloat[]{
    \includegraphics[width=.33\linewidth]{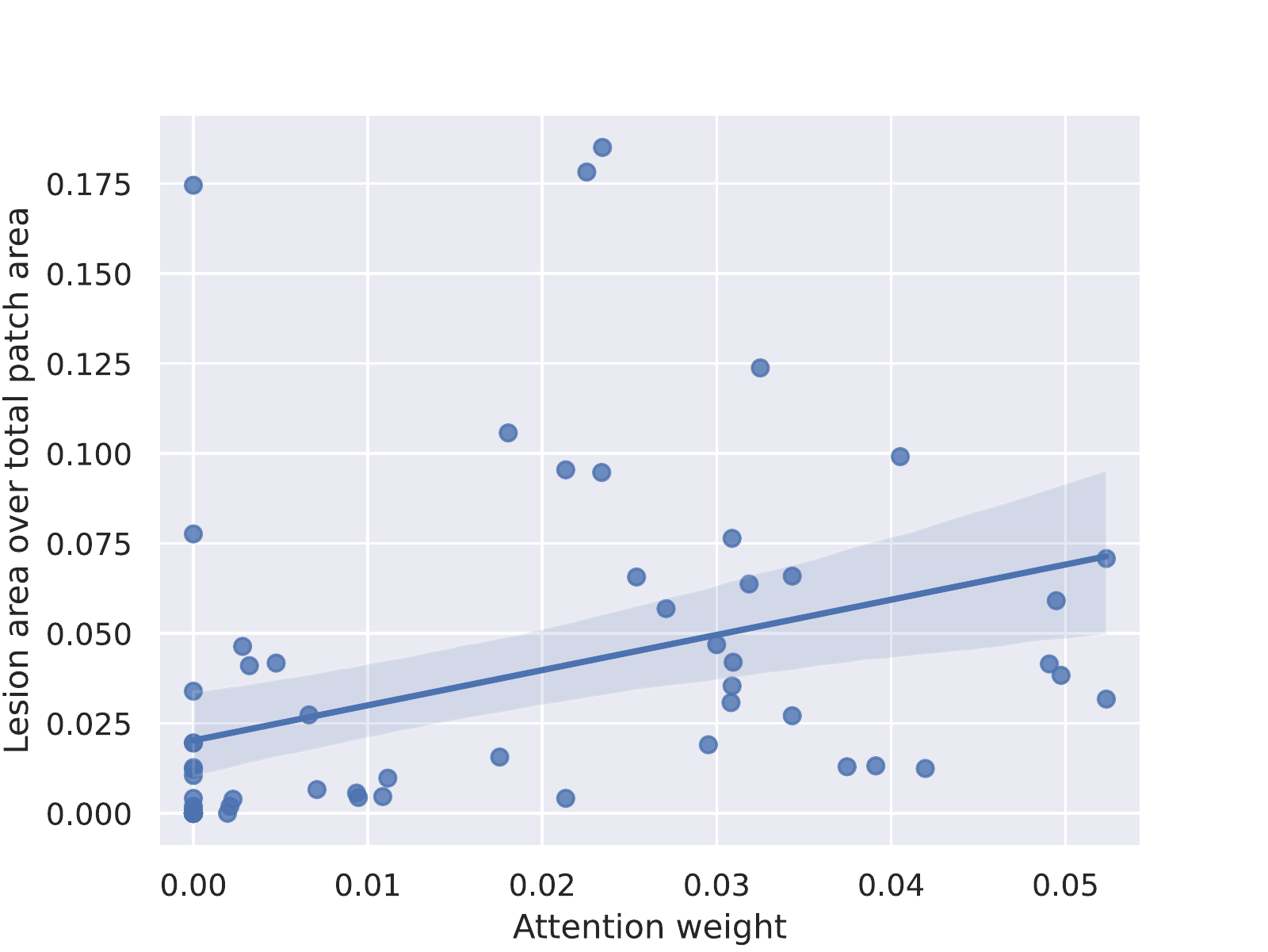}
  }
  \subfloat[]{
    \includegraphics[width=.33\linewidth]{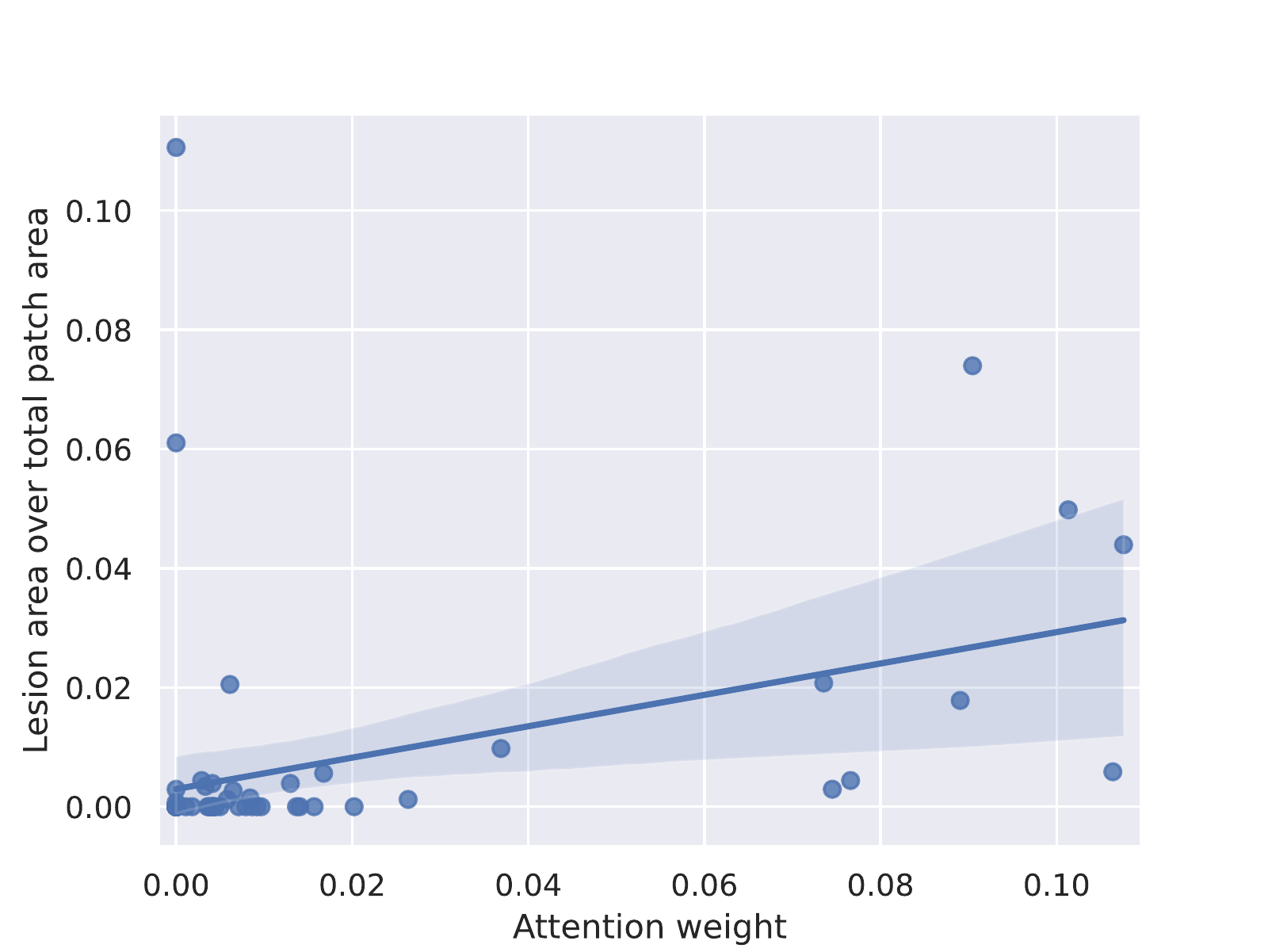}
  }
  \subfloat[]{
    \includegraphics[width=.33\linewidth]{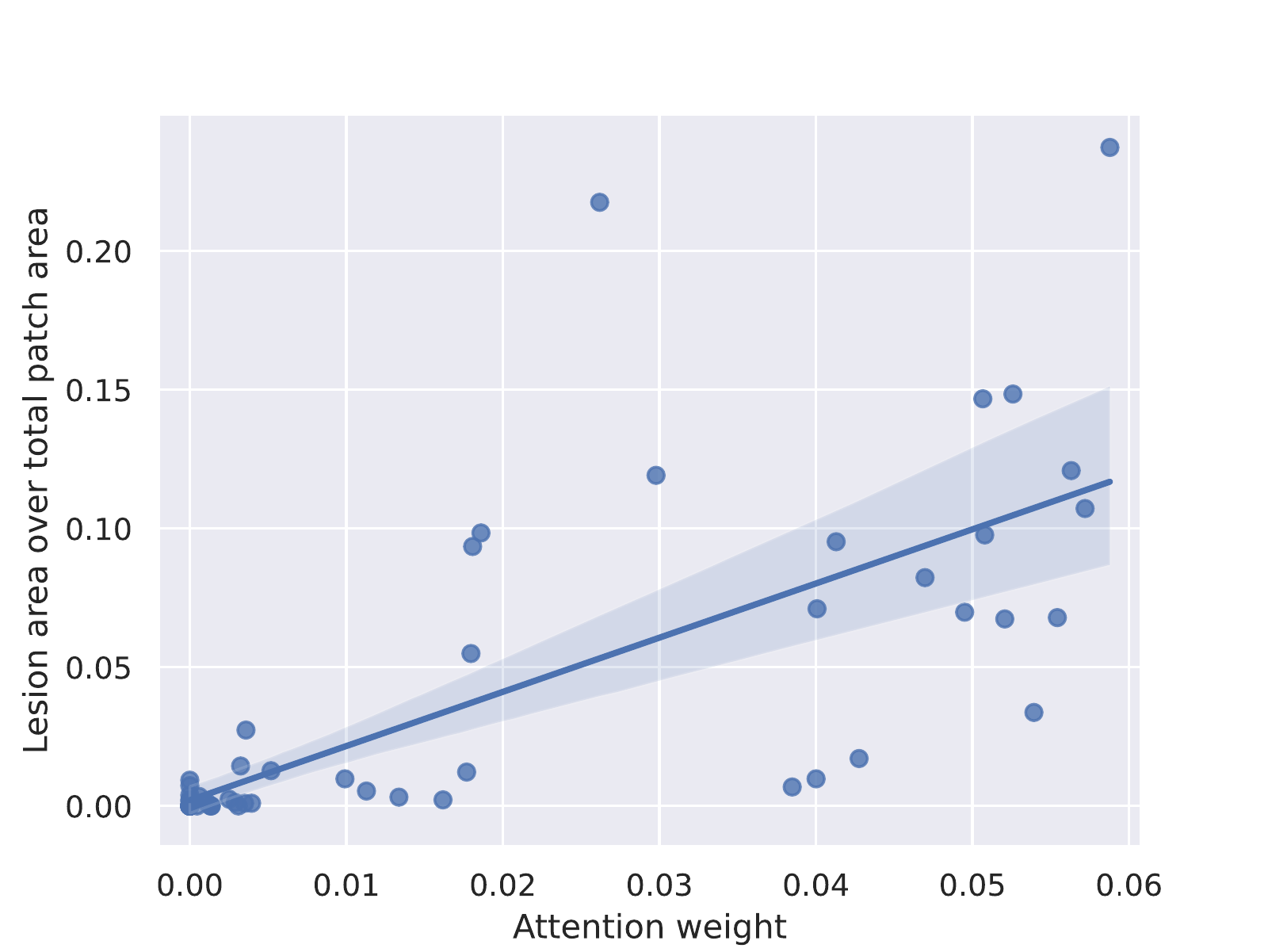}
  }
  \hspace{0mm}



    
  \caption{Example scatterplots of the attention
  weight assigned by the model to each patch versus the 
  percentage of the patch area that contains lesions. The plots
  are computed using images from the IDRiD dataset, for which
  detailed per-pixel lesion annotations for microaneurysms, haemorrhages, soft
  and hard exudates are available. A linear regression fit is estimated for each image and overlayed
  on the plot to highlight the linear trend.} \label{fig:scatterplots}
\end{figure*}

In the second experiment, we are interested in verifying whether 
the attention weight of a patch depends on the amount of lesions
it contains. To that end, we compute the scatterplots
of the attention weights versus the percentage of the area
of a patch that corresponds to lesions (i.e. how many pixels
in a given patch have been annotated as belonging to any
lesion category). We present such plots for 9 images
of the IDRiD dataset in Fig. \ref{fig:scatterplots}.
Based on these figures, we find that there seems to be a 
positive linear correlation between 
the magnitude of the attention weight and the lesion dominance
in the patches of an image. 
Nevertheless, the existence of outliers
suggests that in some cases the size of the lesion may not be
the most dominant factor in the attention distribution, and other factors,
like smaller lesions but with more characteristic shapes, could
be deemed more interesting by the model.

\section{Discussion} \label{sec:discussion}
The development of methods for automated DR grading 
is one of the most popular applications of machine learning
in recent years.
According to a recent survey paper \cite{stolte2020survey},
there is an exponential rise of interest in the field after 2015,
with over 50 papers in 2018 and 70 papers in 2019 using deep
learning. Yet, despite the vast growth, 
there are no standardized benchmarks for 
DR algorithms and, as a result, there is no easy way to compare 
fairly with other works. This is made worse 
by the inherent inter- (and intra-) grader 
variability human experts themselves exhibit \cite{krause2018grader} when
manually grading DR. As different works often
produce their own DR grades for a dataset in cooperation with
their eye specialists, one must keep in mind that even when 
comparing with more standardized datasets such as Messidor-2, 
there might be discrepancies between the DR grades used. 
In this work we made
a conscious effort to facilitate easy comparisons with future work
by employing standard training/test splits and the official DR grades 
associated with each dataset. 

With respect to classification performance, we have demonstrated 
that the attention-based MIL approach 
is a viable alternative to the usual pipelines for rDR detection.
More specifically, it
proved resilient against the noisy setting of the
Kaggle-EyePACS dataset, where, to the best of our knowledge, it achieved
the highest AUC on the single image prediction task.
In the cross-dataset testing scenario, it achieved an AUC
of $0.976$ on Messidor-2, which 
is comparable to the relevant state-of-the-art (AUC of $0.99$)
and often superior to that of other models trained on larger datasets.

Instead of extracting image patches deterministically (i.e. using 
a pre-defined grid), our model was trained
on pools of randomly sampled patches. Naturally, this 
modification greatly improved the training time, as an image
can be represented by as little as $50$ patches, while extracting 
patches over a grid with $50\%$ overlap yields $225$ $64\times64$ patches
for a $512\times512$ image.
Most interestingly, it also slightly increased the model's performance 
on the validation set (Fig. \ref{fig:patch_policy}).
This can be explained by thinking the random patch policy as 
a form of implicit data augmentation: 
each time a specific image is used for training, the model
will, with high probability, see a new bag of patches encoding,
owing to the image being represented by a relatively small
random subset of all possible patches.

Apart from detection accuracy, model 
intrepretability is a common requirement 
in medical applications 
of machine learning, as it 
allows human experts to make sense of why the model makes some
prediction and not the opposite. Thus, interpretability
is key in the acceptance of any proposed machine learning solution
in the field of eye care.
To aid in this, our method outputs a heatmap that
contains the attention weights assigned to each image region 
during inference. This attention heatmap emerges as a natural
by-product of the model and consequently does not require
additional computation, as is the case in alternative 
interpretability approaches for deep networks such
as Grad-cam \cite{selvaraju2017grad}.
Such a heatmap can be used alongside the rDR probability prediction,
to inform medical experts of the particular fundus artifacts that 
led to that prediction. Furthermore, it can help
machine learning practitioners understand their model's failures
and find ways to overcome them. For instance, we can see that in
false positive predictions (images without rDR that are predicted as rDR)
our model usually focuses 
on regions that contain either tiny spots that resemble 
microaneurysms (Fig. \ref{fig:fp3}), yellow lesions that resemble hard
exudates (Fig. \ref{fig:fp2}) or bright cotton-wool-like artifacts (Fig. \ref{fig:fp1}).
Bringing these findings to the attention of eye specialists could facilitate 
a better understanding as to why the model mistakes these 
artifacts for lesions, as well as a potential strategy to 
counter such factors of confusion.

One limitation of the proposed approach is the need for large patch overlap
during inference. While this is not an issue 
during training (due to the random patch sampling), it can pose a memory constraint
on the system's deployment. For classification an overlap value of
$0.75$ (which results in $841$ patches per image) 
achieves good performance and can be handled relatively
easily. Nevertheless, for constructing the auxiliary attention heatmaps
larger values are necessary in order to have more fine-grained 
and pleasing visualizations. As a measure of scale, the images
in Fig. \ref{fig:visualizations} were produced using
a patch overlap of $0.875$, the largest value that a
GPU system with 8GB of memory (NVidia 1070Ti) could handle.

Finally, an interesting line of investigation for future work, would be to question the
utility of using a CNN designed for large resolution images (such as ResNet-18) to
process patches of quite smaller dimensions ($64\times64$). In fact, while
for this work we opted for a clearly over-parameterized approach, it will be interesting
to test whether a model designed to work with smaller resolutions (e.g. 
a ResNet-20 that is designed for $32\times32$ images), performs equally well or even better.
Another topic of future research will be to examine how to utilize 
pixel-level DR lesion annotations, typically available in
very small amounts due to the tediousness of producing them, 
in order to improve the predictive performance 
and lesion heatmap quality. This remains an open problem, as 
DR grading and DR lesion localization methods have remained more or less orthogonal,
with very few published works on the topic \cite{zhou2019collaborative}.

\section{Conclusions} \label{sec:conclusion}
We introduced a system that detects referable Diabetic Retinopathy 
in fundus images, by extracting
local information from each image patch separately and combining it with
an attention mechanism. In our experiments, the 
proposed system achieved high classification performance, making
it competitive with state-of-the-art works that were trained
in larger and better annotated data.
Aside from its high predictive value, our system can inherently produce a heatmap
of the regions on which its decision was based, thus aiding in 
the interpretation of its predictions.

\section*{Acknowledgment}
Messidor-2 data were kindly provided by the Messidor program partners (see http://www.adcis.net/en/third-party/messidor/).


\bibliographystyle{IEEEtran}
\bibliography{root}

\end{document}